\documentclass[10pt,twocolumn,letterpaper]{article}

\usepackage[pagenumbers]{cvpr} 

\usepackage{microtype}

\usepackage{multirow}
\usepackage{pgfplots}
\pgfplotsset{compat=1.18}
\usepackage{adjustbox}
\usepackage{makecell}
\usepackage{booktabs}
\usepackage{colortbl}
\usepackage{graphicx}  
\usepackage{caption}
\usepackage{cuted}
\usepackage{placeins}
\usepackage{float}
\usepackage[accsupp]{axessibility}  

\newcommand{\NAME}{XVR\xspace}

\definecolor{corcolor}{RGB}{255,212,208}   
\definecolor{vercolor}{RGB}{190,255,200}    
\definecolor{loccolor}{RGB}{214,225,255}   
\definecolor{tmpcolor}{RGB}{255,230,173}   

\definecolor{cvprblue}{rgb}{0.21,0.49,0.74}
\usepackage[pagebackref,breaklinks,colorlinks,allcolors=cvprblue]{hyperref}

\usepackage{array}
\usepackage{booktabs}
\usepackage[table]{xcolor}
\usepackage{makecell}
\usepackage{siunitx}
\usepackage{adjustbox}
 
\def\paperID{29050} 
\def\confName{CVPR}
\def\confYear{2026}

\title{Learning Multi-View Spatial Reasoning from Cross-View Relations}

\author{Suchae Jeong$^{*1,2}$ \; Jaehwi Song$^{*2,3}$ \; Haeone Lee$^{1,2}$ \; Hanna Kim$^{1}$ \; Jian Kim$^{4}$ \; Dongjun Lee$^{1}$ \\
Dong Kyu Shin$^{5}$ \, Changyeon Kim$^{1}$ \, Dongyoon Hahm$^{1}$ \, Woogyeol Jin$^{1}$ \, Juheon Choi$^{1}$ \, Kimin Lee$^{1,2}$ \\
{\small $^{1}$KAIST \quad $^{2}$Config \quad $^{3}$Hanyang University \quad $^{4}$Yonsei University \quad $^{5}$Seoul National University} \\
{\small \url{https://cross-view-relations.github.io}}
}

\begin{document}
\twocolumn[{
    \maketitle
    \vspace{-13mm}
    \begin{center}
        \centering
        \captionsetup{type=figure}
        \includegraphics[width=0.95\linewidth]{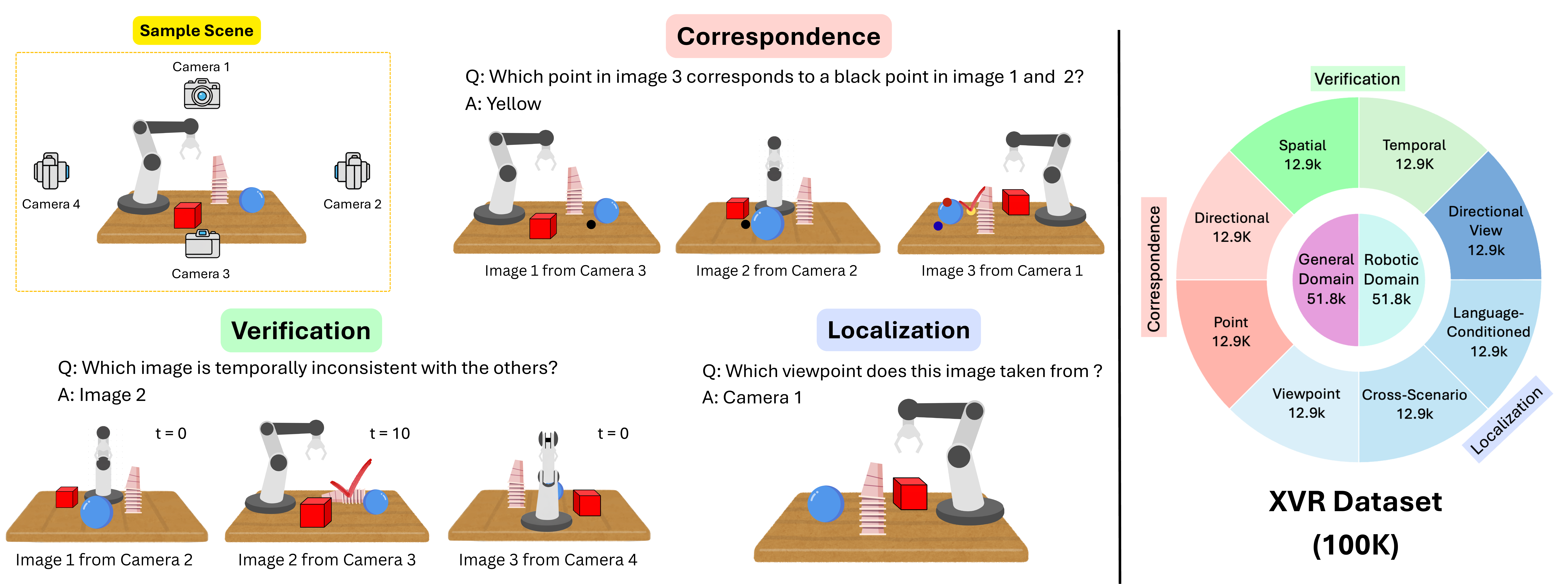}
        \vspace{-0.1in}
        \caption{
        Overview of the Cross-View Relations (XVR). The illustration highlights how multi-view images relate across viewpoints: linking spatial relations (Correspondence), checking cross-view consistency (Verification), and inferring the camera viewpoint (Localization). All XVR dataset samples are derived from real images.}
        \label{fig:teaser}
    \end{center}
}]

\begin{abstract}
Vision-language models (VLMs) have achieved impressive results on single-view vision tasks, but lack the multi-view spatial reasoning capabilities essential for embodied AI systems to understand 3D environments and manipulate objects across different viewpoints. 
In this work, we introduce Cross-View Relations (XVR), a large-scale dataset designed to teach VLMs spatial reasoning across multiple views.
XVR comprises 100K vision-question-answer samples derived from 18K diverse 3D scenes and 70K robotic manipulation trajectories, spanning three fundamental spatial reasoning tasks: Correspondence (matching objects across views), Verification (validating spatial relationships), and Localization (identifying object positions). 
VLMs fine-tuned on XVR achieve substantial improvements on established multi-view and robotic spatial reasoning benchmarks (MindCube and RoboSpatial). 
When integrated as backbones in Vision-Language-Action models, XVR-trained representations improve success rates on RoboCasa. 
Our results demonstrate that explicit training on cross-view spatial relations significantly enhances multi-view reasoning and transfers effectively to real-world robotic manipulation.
\end{abstract}
\section{Introduction}

Vision-Language Models (VLMs) have demonstrated strong performance on visual understanding tasks, such as optical character recognition~\cite{kim2022ocr, lee2023pix2struct, chen2023pali, liu2024llavanext}, image captioning~\cite{radford2021learning, li2023blip, wang2023image}, and video understanding~\cite{bain2021frozen, zellers2021merlot, cheng2024videollama, wang2024internvideo2}. 
Recent work has extended these capabilities to spatial reasoning~\cite{chen2024spatialvlm, cheng2024spatialrgpt, cai2025spatialbot, zha2025enable}, enabling models to reason about object locations, relations, and motion within visual scenes.

\renewcommand{\thefootnote}{\fnsymbol{footnote}}
\footnotetext[1]{Equal contribution}

\begin{table*}[t]
\centering
\renewcommand{\arraystretch}{0.9}
\begin{tabular}{lccllcc}
\toprule
\textbf{Dataset} & \textbf{Split} & \textbf{\# Imgs / sample} & \textbf{Domain} & \textbf{\# Images} & \textbf{\# QAs} \\
\midrule
3DSRBench-real & Eval & 1.00 & General & 2.1K & 2.1K \\
All-Angles-Bench & Eval & 4--5 & General & 450 & 2.1K \\
MMSI-Bench & Eval & 2.55 & General, Robotic & 2K & 1K \\
SpatialVLM & Train, Eval & 1.00 & General & 10M & 2B \\
RoboSpatial & Train, Eval & 1.00 & General & 1M & 3M \\
MindCube & Train, Eval & 3.37 & General & 3.2K & 21K \\
MultiSPA & Train, Eval & 1.85 & General & 1.1M & 27M \\
\midrule
\rowcolor{gray!20} \textbf{XVR (Ours)} & \textbf{Train, Eval} & \textbf{4.32} & \textbf{General, Robotic} & \textbf{447K} & \textbf{103K} \\
\bottomrule
\end{tabular}
\caption{Comparison of spatial reasoning datasets. XVR provides the highest mean images per sample among training datasets, with supervision spanning both general and robotic domains.}
\label{tab:dataset_comparison}
\end{table*}

However, existing spatial reasoning research has focused almost exclusively on single-view settings. Most VQA datasets and spatial reasoning benchmarks~\cite{liu2023visual, zhang2024vision, ma20253dsrbench, du2024embspatial, jia2025omnispatial, chen2024spatialvlm, shiri2024empirical, song2025robospatial} provide only a single viewpoint, which suffers from limited spatial information and frequent occlusions. This is particularly problematic given that multi-camera setups have become standard in robotics applications~\cite{BerkeleyUR5Website, shi2023robocook, ConqHoseManipData, sawhney2020playing, luo2307multi, matsushima2023weblab, dasari2019robonet, khazatsky2024droid, luo2025fmb, wang2023mimicplay, fu2024mobile, kumar2023robohive}, where understanding geometric relationships between viewpoints is essential for tasks such as manipulation and navigation. While recent work has introduced multi-view datasets~\cite{yeh2025seeing, yin2025spatial, feng2025seeing}, these focus primarily on identifying what objects appear in each view, rather than understanding how different viewpoints relate geometrically. Without explicit supervision on cross-view spatial relationships, VLMs often generate predictions that appear visually plausible within individual views but are spatially inconsistent across viewpoints.


To address this limitation, we introduce Cross-View Relations (XVR), a dataset of 100K multi-view VQA samples that provides explicit supervision on geometric relationships across viewpoints.
Drawing inspiration from Structure-from-Motion (SfM)~\cite{ozyecsil2017survey, schonberger2016structure}, we design three reasoning primitives that capture how views relate geometrically: (i) Cross-view Correspondence: identifying matching elements across views, (ii) Geometric Consistency Verification: validating whether view relationships are geometrically plausible, and (iii) Relative Viewpoint Localization: reasoning about spatial relationships between camera perspectives (see Figure~\ref{fig:teaser}).

To construct XVR at scale, we leverage two complementary data sources. Calibrated multi-view captures (the general domain) provide dense geometric supervision with accurate camera parameters, enabling precise correspondence and consistency annotations. Robotic trajectories (the robotic domain) contribute temporal continuity and diverse viewpoint transitions from embodied interactions, enriching the dataset with dynamic perspective changes. Together, these sources provide the geometric precision and viewpoint diversity needed for comprehensive cross-view reasoning.

    Evaluation across ten VLMs (both open-source~\cite{li2025eagle, beyer2024paligemma, wang2025internvl3, yang2025qwen3} and closed-source  models~\cite{claude4.5, gpt5, comanici2025gemini, abdolmaleki2025gemini}) demonstrates substantial improvements: models trained with XVR achieve a 1.8× relative gain in accuracy on XVR-Eval (our internal benchmark) and show consistent improvements on external benchmarks including MindCube-Tiny~\cite{yin2025spatial} and RoboSpatial-Home~\cite{song2025robospatial}. Furthermore, when XVR-trained VLMs serve as backbones for Vision-Language-Action (VLA) models, they yield significant gains, improving manipulation success rates on simulated environments from RoboCasa~\cite{nasiriany2024robocasa} by an average of 13\% absolute. This demonstrates that cross-view relation reasoning transfers effectively to real-world robotic control.

\vspace{15pt}
Our contributions are summarized as follows:
\begin{itemize}
\item We introduce XVR, a dataset with explicit supervision on cross-view relations for multi-view spatial reasoning.
\item XVR contains 100K samples spanning two complementary domains, i.e., general scenes and robotic trajectories, organized into three task categories (Correspondence, Verification, and Localization) across eight specific tasks.
\item We show that training on XVR improves performance on XVR-Eval, transfers to external multi-view and robotic spatial benchmarks, and enhances downstream VLA manipulation performance.
\end{itemize}

\section{Related Work}

\paragraph{Single-view Spatial Reasoning}
Spatial reasoning research has primarily focused on single-view settings. Early work established baselines on synthetic scenes~\citep{johnson2017clevr} and extended them to real images with relational structure~\citep{hudson2019gqa}. Subsequent studies exposed failures in directional reasoning~\citep{liu2023visual}, distance estimation~\citep{shiri2024empirical}, and frame-of-reference understanding~\citep{du2024embspatial, jia2025omnispatial}.
To address these limitations, recent methods inject 3D cues through large-scale supervision~\citep{chen2024spatialvlm}, augment features with depth and scene structure~\citep{cheng2024spatialrgpt}, or simulate viewpoint changes via abstract 3D proxies~\citep{lee2025perspective}. However, single-view observations provide limited spatial information and often suffer from occlusions. This motivates multi-view approaches where cross-view relations become essential.

\paragraph{Multi-view Spatial Reasoning}
Multi-view settings address single-view limitations by leveraging complementary viewpoints. Prior work transfers knowledge across views for improved QA~\citep{zhang2025open3dvqa, ma2022sqa3d} and probes viewpoint robustness through relative direction, distance, and 6D pose~\citep{li2024mvbench, hong20233d, mo2025advancing}.
Recent benchmarks evaluate multi-view understanding across diverse settings. AllAnglesBench~\citep{yeh2025seeing} tests perspective-taking abilities. MindCube~\citep{yin2025spatial} assesses spatial reasoning from limited views. 3DSRBench~\citep{ma20253dsrbench} probes viewpoint robustness by varying camera poses. These benchmarks primarily focus on object properties within views or object-view grounding rather than cross-view relations. Large-scale datasets with explicit cross-view supervision remain limited. Recent works have made progress in multi-frame spatial reasoning: MultiSPA~\cite{xu2025multi} provides large-scale training data for depth and visual correspondence, and MMSI-Bench~\cite{yang2025mmsi} offers a human-curated evaluation benchmark for multi-image spatial intelligence. However, these works either lack explicit supervision on cross-view geometric relationships or do not cover both general and robotic domains. XVR addresses this gap by providing dense cross-view supervision across both domains, with an average of 4.32 images per sample.


\paragraph{Vision-Language-Action Models}
Recent VLA models map vision-language inputs directly to actions~\citep{zitkovich2023rt, kim2024openvla, black2410pi0, o2024open, bjorck2025gr00t, shi2025hi}. To enhance spatial reasoning in VLA backbones, recent work injects robot-specific spatial signals~\citep{song2025robospatial, feng2025seeing} and develops trajectory-grounded QA~\citep{sermanet2024robovqa, chen2025robo2vlm, ji2025robobrain}. Methods like pi0.5~\citep{intelligence2025pi_} demonstrate improved embodied reasoning through enhanced VLM backbones.
XVR leverages robotic trajectories to construct datasets with explicit cross-view relation supervision. VLMs trained with XVR serve as improved backbones for VLA models, enhancing embodied manipulation performance.
\begin{figure*}[t] 
  \centering
  \includegraphics[width=0.95\textwidth]{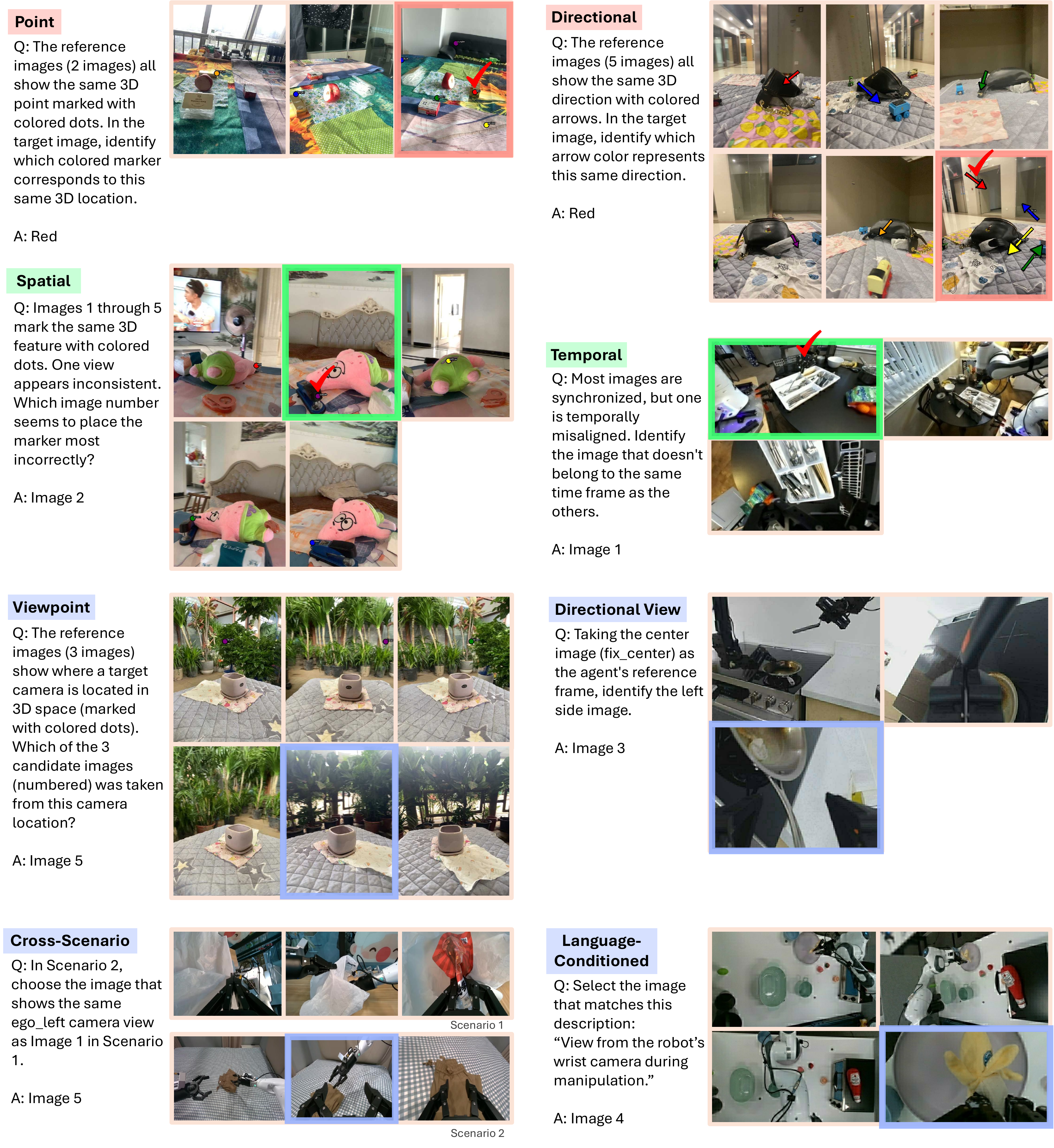}
  \caption{
    \textbf{Overview of the question–answer (QA) structure in \NAME.}
    The figure shows representative examples from eight task types across correspondence, verification, and localization categories, demonstrating the consistent QA format used throughout the dataset. Each category is color-coded: red for Correspondence (Point, Directional), green for Verification (Spatial, Temporal), and blue for Localization (Viewpoint, Directional View, Cross-Scenario, Language-Conditioned). 
  }
  \label{fig:2_QA_representative_overview}
\end{figure*}

\section{Cross-View Relation Dataset}
\label{sec:dataset}
We introduce Cross-View Relation (XVR), a dataset 
for learning multi-view spatial reasoning through explicit 
cross-view relation supervision. 

\subsection{Task Categories}
\label{sec:tasks}
Multi-view spatial reasoning requires understanding how different viewpoints relate to each other geometrically. We organize XVR into the following three task categories: 

\begin{itemize}
    \item \textbf{Correspondence}: Identifying matching elements across views that represent the same physical entity. Tasks in this category teach models to link visual features across different viewpoints, forming the foundation for understanding shared scene geometry across views.
    
    \item \textbf{Verification}: Checking whether multi-view observations are geometrically or temporally consistent. Tasks in this category teach models to detect spatial or temporal inconsistencies, ensuring their understanding maintains coherence across views.
    
    \item \textbf{Localization}: Determining relative camera positions and which viewpoint corresponds to specific spatial conditions. This category captures how cameras relate to each other spatially and enables reasoning about relative viewpoints.
\end{itemize}

Together, these three categories provide structured supervision for learning cross-view relations, enabling robust multi-view spatial reasoning. We operationalize them through eight tasks. Figure~\ref{fig:2_QA_representative_overview} illustrates the three categories with representative examples.

\paragraph{Connection to Structure-from-Motion.} Our categorization draws inspiration from Structure-from-Motion (SfM)~\cite{ozyecsil2017survey, schonberger2016structure}, a classical approach that integrates geometric information across multiple views to reconstruct 3D scenes. SfM operates through three key stages that directly inspired our categories: (i) identifying correspondences across views, (ii) verifying geometric consistency, and (iii) estimating camera poses. We adapt these stages into cross-view supervision for multi-view spatial reasoning.

\subsection{Task Definitions}
We instantiate the three categories through eight specific tasks.

\paragraph{Correspondence.}
\textit{Point Correspondence} requires identifying which point across multiple views represents the same physical location in 3D space. This task evaluates whether models can match spatially aligned visual features under viewpoint changes.
\textit{Directional Correspondence} extends this to 3D orientation, requiring models to align directional arrows or vectors consistently across different camera projections. It tests reasoning about directional geometry beyond simple point matching.

\paragraph{Verification.}
\textit{Spatial Verification} requires detecting correspondences that violate 3D spatial consistency among multiple views. By identifying geometrically inconsistent matches, this task measures the model's ability to enforce spatial coherence across perspectives. 
\textit{Temporal Verification} requires identifying temporally inconsistent frames within a sequence. It assesses understanding of spatial-temporal structure by detecting frames that break temporal continuity.

\paragraph{Localization.}
\textit{Viewpoint Localization} determines which camera view corresponds to a specific spatial position in the scene. This task evaluates whether models can infer relative viewpoint positions based on visual cues from multiple reference views.
\textit{Directional View Localization} identifies which camera view is located in a specific direction (e.g., left or right) relative to a reference camera. It evaluates directional awareness and relational reasoning between viewpoints.
\textit{Cross-Scenario Localization} requires matching corresponding viewpoints across structurally similar but distinct scenes. This task examines the generalization of viewpoint reasoning under scene-level variations.
\textit{Language-Conditioned Localization} selects the camera view that best matches a natural language spatial description. It integrates linguistic spatial cues (e.g., wrist-mounted camera) with geometric reasoning to identify corresponding visual perspectives.

\subsection{Data Generation Pipeline}
\label{sec:generation}

To instantiate the eight tasks, we develop a unified generation framework (denoted as $\mathcal{G}$). 
This framework operationalizes our cross-view relation categories by structuring raw multi-view data to concrete question-answer (QA) pairs. 
As formalized in the supplementary material (Eq.~\ref{eq:qa_framework}), our framework is defined as $\mathcal{G}:(\mathcal{I},\mathcal{P},X,\mathcal{T},\mathcal{M})\rightarrow(\mathcal{Q},\mathcal{A})$, where inputs comprise images ($\mathcal{I}$), camera parameters ($\mathcal{P}$), 3D geometry ($X$), temporal indices ($\mathcal{T}$), and metadata ($\mathcal{M}$).

The generation process differs based on data source characteristics. We describe two primary pipelines: the general domain pipeline, which leverages explicit 3D geometric information, and the robotic domain pipeline, which utilizes spatio-temporal metadata from robotic trajectories.


\paragraph{General domain.} 
For tasks leveraging explicit 3D geometry (\textit{Point Correspondence}, \textit{Directional Correspondence}, \textit{Spatial Verification}, and \textit{Viewpoint Localization}), we employ a 3D-to-2D projection approach. We sample 3D primitives—points for correspondence tasks, camera positions for localization tasks—that are visible across multiple views. Using camera parameters from $\mathcal{P}$, we project these primitives onto available views and construct reference-target QA pairs. To create challenging questions, we generate spatially separated distractors for multiple-choice options, ensuring models must perform genuine cross-view reasoning rather than relying on low-level visual cues.

\paragraph{Robotic domain.} 
For tasks utilizing robotic trajectories (\textit{Temporal Verification}, \textit{Directional View Localization}, \textit{Cross-Scenario Localization}, and \textit{Language-Conditioned Localization}), we sample from spatio-temporal metadata $\mathcal{M}$ and temporal indices $\mathcal{T}$. A critical quality control step ensures generated questions are perceptually meaningful: for Temporal Verification, we employ SSIM-based filtering~\cite{wang2004image} combined with action-based heuristics to verify that temporal differences produce visually distinguishable scene changes. This filtering prevents trivial questions where images are perceptually identical despite different timestamps.

All tasks follow a consistent reference-target QA structure where multiple reference views provide context and models must identify correct answers through cross-view reasoning. Complete task formalization is provided in Table~\ref{tab:qa-task}. Further details on the generation pipeline are provided in Appendix~\ref{app:generation_pipeline} with an illustration in Figure~\ref{fig:generation_pipeline}.

\subsection{Data Sources and Curation}
\label{sec:data_sources}

We construct XVR using the following specific sources. These sources provide geometric richness from calibrated multi-view captures and realistic embodied dynamics from robotic trajectories, forming a balanced foundation for multi-view spatial reasoning.

\paragraph{General Domain.}
General domain data provides dense geometric supervision with accurate camera calibration, essential for geometry-based task generation.
We adopt WildRGB-D~\cite{xia2024rgbd} as our primary source, which contains multi-view RGB-D captures of diverse scenes with calibrated camera parameters.
To ensure reliable geometric grounding and high-quality QA generation, we retain only samples with sufficiently dense point clouds (at least 1M points), guaranteeing robust 3D-to-2D projection and visibility analysis.

\paragraph{Robotic Domain.}
Robotic domain data provides temporal continuity and viewpoint diversity observed during manipulation tasks.
We leverage OXE~\cite{o2024open} and AgiBot-World~\cite{bu2025agibot} datasets as primary sources.
Given the variable quality in raw robotic data, we apply strict filtering criteria to ensure task validity:
(1) We include only sequences providing at least three distinct camera views to enable meaningful multi-view reasoning. Among publicly available datasets within the OXE suite, only DROID~\cite{khazatsky2024droid}, MobileAloha~\cite{fu2024mobile}, RoboSet~\cite{kumar2023robohive}, and FMB~\cite{luo2025fmb} satisfy this requirement.
(2) We exclude sequences with inconsistent or ambiguous camera identifiers, as these compromise metadata-based localization task accuracy.
(3) We retain only trajectories lasting at least 20 seconds with sufficient motion dynamics, measured by end-effector displacement, ensuring perceptually meaningful temporal variations for verification tasks. Further details on data sources and distribution are provided in Appendix~\ref{app:dataset_statistics}.

\begin{table*}[t]
    \centering
    \footnotesize
    \setlength{\tabcolsep}{3pt}

    \begin{adjustbox}{max width=\textwidth}
    \begin{tabular}{@{}l
                    S[table-format=2.2, table-parse-only]
                    S[table-format=2.2, table-parse-only]
                    S[table-format=2.2, table-parse-only]
                    S[table-format=2.2, table-parse-only]
                    S[table-format=2.2, table-parse-only]
                    S[table-format=2.2, table-parse-only]
                    S[table-format=2.2, table-parse-only]
                    S[table-format=2.2, table-parse-only]
                    S[table-format=2.2, table-parse-only]
                    S[table-format=2.2, table-parse-only]
                    @{}} 

    \toprule
    & \multicolumn{2}{c}{\cellcolor{corcolor}\textbf{Correspondence}} 
      & \multicolumn{2}{c}{\cellcolor{vercolor}\textbf{Verification}} 
      & \multicolumn{4}{c}{\cellcolor{loccolor}\textbf{Localization}} & \\
    \cmidrule(lr){2-3}\cmidrule(lr){4-5}\cmidrule(l){6-9}
    \textbf{Model}  
        & {\textbf{Point}} & {\textbf{Directional}}
        & {\textbf{Spatial}} & {\textbf{Temporal}}
        & {\textbf{Viewpoint}} & {\makecell{\textbf{Directional}\\\textbf{View}}} & {\makecell{\textbf{Cross-}\\\textbf{scenario}}} & {\makecell{\textbf{Language-}\\\textbf{conditioned}}} & {\textbf{Overall}}\\
    \midrule
    \rowcolor{tmpcolor}
    \multicolumn{10}{@{}l}{\textit{Closed-source Models}}\\
    Claude-4.5-Sonnet & 68.94 & 24.24 & 52.65 & 51.76 & 23.65 & \textbf{63.35} &	71.95 & 57.01 & 51.18\\
    GPT-5 &	\textbf{83.33} &	\textbf{32.20} &	\textbf{68.56} &	\textbf{65.29} &	\textbf{38.59} &	60.63 &	\textbf{80.54} &	\textbf{67.87} & \textbf{61.74}\\
    Gemini-2.5-flash & 78.03 & 31.44 & 60.61 & 56.47 & 14.52 & 57.47 & 66.06 & 56.11 & 52.36\\
    Gemini-2.5-Pro & 74.24 & 26.14 & 56.06 & 50.59 & 24.48 & 52.94 & 60.18 & 48.42 & 49.04\\
    Gemini-Robotics-ER-1.5 & 76.89 & 22.35 & 50.00 & 51.76 & 6.22 & 53.85 & 66.06 & 56.11 & 47.48\\
    \midrule
    \rowcolor{tmpcolor}
    \multicolumn{10}{@{}l}{\textit{Open-source Models}}\\
    Eagle2-2B &	20.45 &	23.86 &	20.08 &	31.18 &	0.41 &	14.48 &	27.60 &	0.00 & 16.99\\
    paligemma2-3b &	2.65 &	4.55 &	23.11 &	35.29 &	6.64 &	30.77 &	11.76 &	33.48 & 17.36 \\
    InternVL-3.5-4B & 34.09 &	25.00 &	24.62 &	49.41 &	4.15 &	52.04 &	37.10 &	41.18 & 32.32\\
    Qwen3-VL-2B-Instruct &	46.59 &	26.14 &	23.11 &	45.29  &	19.50 &	47.06 &	41.63 &	51.58 & 36.82 \\
    Qwen3-VL-4B-Instruct  &	57.95 &	29.55 &	48.11 &	\textbf{51.76} &	10.37 &	53.39 &	60.63 &	52.94 & 45.02\\
    Qwen3-VL-2B-XVR (Ours)  &	\textbf{94.32} &	\textbf{53.79} &	\textbf{84.85} &	41.18 &	\textbf{57.68} &	\textbf{68.33} &	\textbf{70.14} &	\textbf{63.35} & \textbf{68.06}\\
    \midrule
    \rowcolor{tmpcolor}
    \multicolumn{10}{@{}l}{\textit{Baseline}}\\
    Random & 20.00 & 25.00 & 22.22 & 33.33 & 33.33 & 50.00 & 33.33 & 50.00 & 32.64\\
    Human & \textbf{92.31} & \textbf{67.11} & \textbf{88.46} & \textbf{77.08} & \textbf{64.94} & \textbf{92.08} & \textbf{87.74} & \textbf{93.48} & \textbf{83.85}\\
    \bottomrule
    \end{tabular}
    \end{adjustbox}
    \caption{Performance comparison on \textbf{XVR-Eval} (\%). Results include closed-source models, open-source models (zero-shot and + XVR), and baselines.}
    \captionsetup{font=footnotesize}

    \label{tab:sft-bench-results}
\end{table*}

\section{Experiments}
\label{sec:experiments}

We conduct three complementary experiments to thoroughly evaluate the impact of XVR on multi-view spatial reasoning. First, we benchmark models on our proposed XVR-Eval suite (Sec.~\ref{sec:XVR_eval}). Second, we evaluate models on external spatial benchmarks (Sec.~\ref{sec:ood_generalization}). Finally, we examine embodied transfer by integrating XVR-trained backbones into a Vision-Language-Action (VLA) model (Sec.~\ref{sec:embodied_transfer}).

\subsection{Benchmarking on XVR-Eval}
\label{sec:XVR_eval}

\paragraph{Setup.} To evaluate cross-view relation reasoning, we construct XVR-Eval, which consists of 1,866 held-out samples constructed from data sources unseen during XVR creation. Specifically, we include new sources: MobileAloha trajectories and WildRGB-D boat category scenes in XVR-Eval. We refer readers to Appendix~\ref{app:XVR_eval_details} for statistics of XVR-Eval.


Using XVR-Eval, we test both open-source VLMs, such as Eagle2-2B~\cite{li2025eagle}, Paligemma-3B~\cite{beyer2024paligemma}, InternVL-3.5-4B~\cite{wang2025internvl3}, and Qwen3-VL-Instruct (2B and 4B variants)\cite{yang2025qwen3}, and closed models: Claude-4.5-Sonnet\cite{claude4.5}, GPT-5~\cite{gpt5}, Gemini-2.5-Flash, Gemini-2.5-Pro~\cite{comanici2025gemini}, and Gemini-Robotics-ER-1.5~\cite{abdolmaleki2025gemini}.
To verify the benefits of our XVR dataset, we fine-tune Qwen3-VL-Instruct (2B) on our XVR dataset and denote it as Qwen3-VL-2B-XVR. We also report a human baseline established from nine researchers with at least four years of higher education, collecting 795 annotations across all tasks.


\paragraph{Main results.}
Table~\ref{tab:sft-bench-results} shows that most open-source models perform near chance level, while closed-source models achieve substantially higher performance yet still fall short of human baselines, indicating significant room for improvement. Our model, Qwen3-VL-2B-XVR, achieves a 1.8$\times$ improvement over its base model and ranks first among all evaluated models, surpassing both open-source and closed-source alternatives. Notably, Qwen3-VL-2B-XVR exceeds human performance on Point Correspondence, demonstrating that targeted supervision on cross-view relations can substantially improve spatial reasoning capabilities.


\paragraph{Task-specific patterns.}
Our analysis reveals two key findings. First, geometric reasoning tasks benefit substantially from XVR training. Point Correspondence and Spatial Verification show dramatic improvements, with Spatial Verification surpassing even GPT-5. Localization tasks demonstrate consistent gains, with Viewpoint Localization approaching human-level performance. These results validate that cross-view supervision enables models to perform geometric consistency checking, precise point matching, and camera-relative reasoning.

Second, Temporal Verification declines after XVR training, the only task showing this pattern. This reveals a trade-off: since most XVR tasks emphasize spatial reasoning at synchronized time points, training biases the model toward geometric structure at the expense of temporal sensitivity.

\paragraph{Closed-source model analysis.}
Despite their scale, closed-source models reveal task-specific limitations. GPT-5 exhibits large within-category variance: it excels at Point Correspondence but struggles with Directional Correspondence, despite both testing correspondence reasoning. Similarly, GPT-5 handles Spatial Verification well but fails at Viewpoint Localization. 

Gemini-Robotics-ER-1.5 achieves the lowest accuracy among closed-source models. Its Viewpoint Localization accuracy (6.22\%) falls below random guessing (22.22\%), indicating minimal camera-relative reasoning capability. Even robotics-specialized training does not develop view-view relation reasoning without explicit supervision.

Gemini-2.5-Flash outperforms Gemini-2.5-Pro despite smaller scale. This shows that model capacity alone does not improve spatial reasoning. After XVR training, Qwen3-VL-2B surpasses all closed-source models, demonstrating that explicit supervision on view relations outweighs scale.

\paragraph{Human baseline comparison.}
XVR-trained models achieve super-human performance on Point Correspondence and Spatial Verification. However, gaps remain on Directional Correspondence and Temporal Verification, where human performance exceeds model performance by over 10 and 35 percentage points, respectively. Models excel at precise geometric calculations while humans handle ambiguous orientations and temporal dynamics better.

\begin{figure*}[t]
  \centering
  \includegraphics[width=\linewidth]{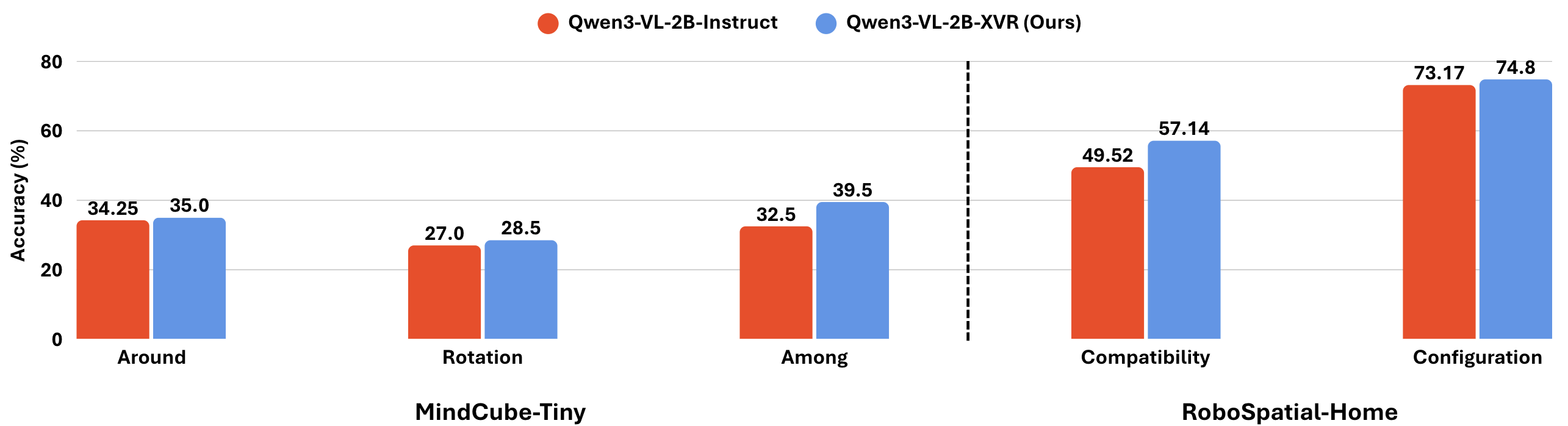}
  \caption{Generalization to external spatial benchmarks (MindCube-Tiny and RoboSpatial-Home). 
Training on XVR improves Qwen3-VL-2B across all tasks, with the largest gains in Compatibility (+7.6\%) and Among (+7.0\%).}
  \label{fig:ood_results}
\end{figure*}

\subsection{Evaluation on External Benchmarks}
\label{sec:ood_generalization}

We test on two external benchmarks not used during XVR creation. MindCube-Tiny~\cite{yin2025spatial} evaluates scene imagination from limited viewpoints through three subtasks: \textit{Around} (object identification under assumed camera motion), \textit{Rotation} (spatial understanding from 360-degree viewpoints), and \textit{Among} (object localization from alternative camera views). RoboSpatial-Home~\cite{song2025robospatial} evaluates spatial understanding for robotic manipulation through three subtasks, of which we evaluate two: \textit{Compatibility} (spatial fit assessment) and \textit{Configuration} (object-object spatial relations). We exclude the Context subtask as all evaluated models score 0. We compare baseline Qwen3-VL-2B against the XVR-trained variant.

Figure~\ref{fig:ood_results} shows that XVR training improves performance across subtasks in both benchmarks, though improvement magnitude varies systematically across tasks.

\paragraph{Transfer patterns.}
Tasks aligned with XVR's training distribution show substantial improvements. MindCube Among requires object localization from alternative camera views, directly matching XVR's multi-view training. RoboSpatial Compatibility and Configuration improve despite testing object-object spatial reasoning, suggesting that cross-view relation training builds 3D representations that transfer more broadly.

Tasks requiring camera motion understanding show minimal improvements. MindCube Around and Rotation involve continuous camera movement patterns absent from XVR's training distribution. XVR consists of 50\% static multi-view scenes and 50\% robotic trajectories that emphasize static camera configurations during manipulation. The limited transfer to motion-based tasks aligns with our temporal reasoning limitations on XVR-Eval.

\paragraph{Distribution shift.}
The improvements occur despite substantial distribution shifts. MindCube uses outside-looking-inward camera configurations, absent from XVR training data which focuses on inside-looking-outward setups. RoboSpatial evaluates single-view spatial reasoning while XVR trains on multi-view relations. These cross-domain improvements validate that cross-view relation reasoning captures general spatial principles rather than dataset-specific patterns.

Despite training exclusively on cross-view relation tasks, XVR-trained models show improvements on object-object spatial reasoning and partially on object-view reasoning across external benchmarks. This demonstrates that cross-view relation supervision provides a foundation for certain aspects of broader spatial reasoning, particularly those involving geometric relationships. Detailed task-by-task analysis is provided in Appendix~\ref{app:ood_analysis}.

\begin{figure}[t]
  \centering
  \includegraphics[width=\columnwidth]{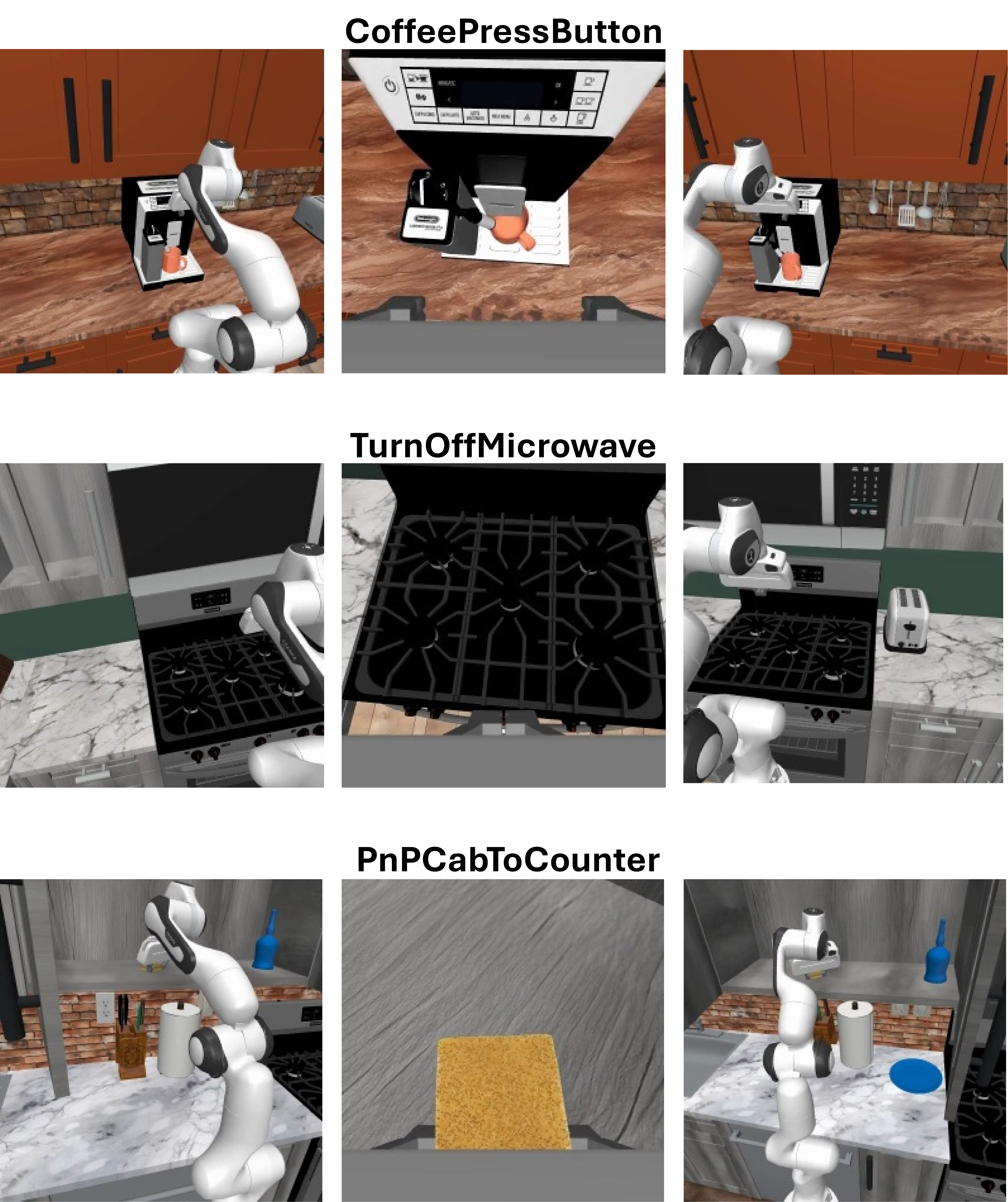}
    \caption{
    Visualization of the three manipulation tasks and their camera-view configurations used for VLA transfer evaluation.
    }
  \label{fig:VLA_transfer_evaluation_tasks}
\end{figure}

\subsection{Transfer to Vision-Language-Action Models} 
\label{sec:embodied_transfer}

To investigate the benefits of XVR on embodied tasks, we extend VLMs trained on XVR into Vision-Language-Action (VLA) models.
Specifically, we add a diffusion action head to VLM representations following the architecture design of GR00T-N1.5 VLA~\cite{bjorck2025gr00t}.
Using the NVIDIA GR00T-X-Embodiment-Sim dataset from the RoboCasa simulator~\cite{nasiriany2024robocasa}, we train VLAs to control a Franka Emika arm performing various manipulation tasks.
We compare a VLA model based on Qwen3-VL-2B-Instruct against one based on our VLM, Qwen3-VL-2B-XVR, and report average success rates across 1,000 rollouts.


We evaluate three manipulation scenarios that require different forms of cross-view spatial reasoning. \textit{CoffeePressButton} involves locating and pressing a small button that is visible only from the wrist camera due to occlusion, testing precise relative distance estimation under partial observability. \textit{TurnOffMicrowave} presents the opposite visibility pattern—the control panel is clearly observed from the left and right cameras but occluded from the wrist view—requiring spatial disambiguation among multiple similar buttons across complementary viewpoints. \textit{PnPCabToCounter} requires grasping one of 64 randomly selected object categories and placing it on the counter, testing generalizable multi-view pose estimation across diverse objects.

\begin{figure}[t]
  \centering
  \includegraphics[width=\columnwidth]{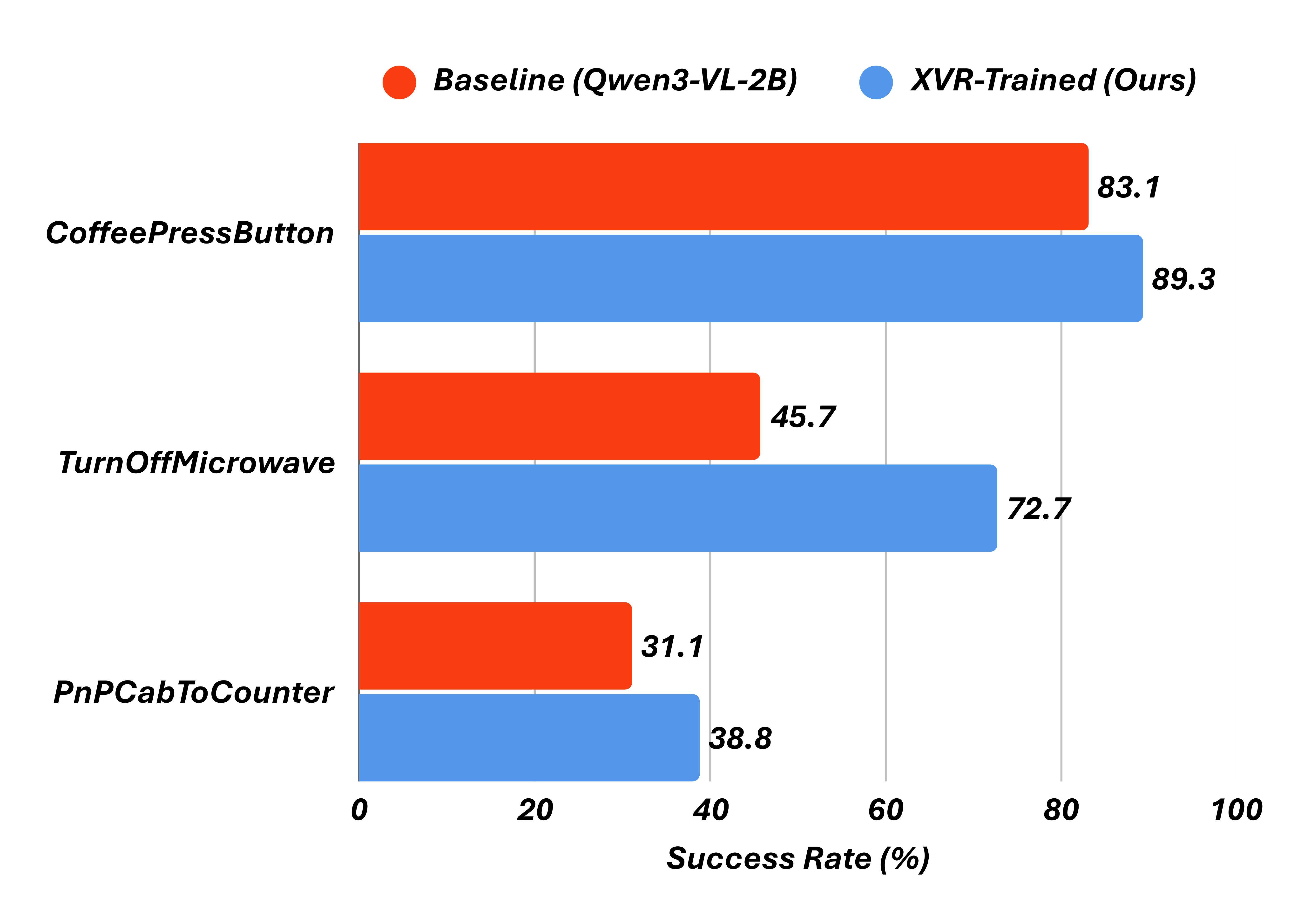}
    \caption{
      \textbf{Transfer to Embodied Tasks: RoboCasa VLA Performance.}
      Fine-tuning on XVR improves Qwen3-VL-2B performance on RoboCasa manipulation tasks, 
      showing effective transfer of spatial reasoning skills to robotic action prediction.
    }
  \label{fig:robocasa_vla_performance}
\end{figure}

Figure~\ref{fig:robocasa_vla_performance} shows that our models consistently improve manipulation performance across all three tasks, with the largest gains on \textit{TurnOffMicrowave}, where cross-view spatial disambiguation is most critical.

These improvements arise from the specific cross-view relation capabilities learned during XVR fine-tuning. Correspondence tasks teach point-level alignment across views, enabling view-consistent 3D representations that support accurate relative distance estimation. Localization tasks provide explicit camera-pose understanding, improving the integration of complementary viewpoints under partial observability. Verification tasks strengthen geometric consistency checking across views, supporting robust pose estimation for diverse object categories. The substantial gains on tasks requiring partial observability, spatial disambiguation, and cross-view generalization demonstrate that cross-view relation supervision enhances the geometric understanding necessary for downstream VLA manipulation.

\section{Conclusion}
\label{sec:conclusion}

We introduce XVR, a dataset for learning multi-view spatial reasoning from cross-view relations. Unlike existing multi-view datasets that emphasize objects within individual views, XVR provides explicit supervision on geometric relationships between views themselves. XVR comprises 100k samples from calibrated multi-view captures and robotic trajectories, organized into three reasoning categories: Correspondence, Verification, and Localization. We also introduce XVR-Eval, a 1,866-sample benchmark for systematic evaluation.
Models trained on XVR demonstrate substantial improvements on XVR-Eval and consistent gains on external multi-view and robotic spatial benchmarks. When integrated into Vision-Language-Action models, XVR-trained backbones improve manipulation success rates on embodied tasks. These results demonstrate that explicit supervision on cross-view relations enhances multi-view spatial reasoning and transfers effectively to embodied manipulation.

This work enables more robust perception for robotic systems that rely on multi-camera setups. Beyond robotics, the approach has broader implications for applications requiring spatial understanding across multiple viewpoints, including autonomous navigation and AR/VR systems where maintaining geometric consistency is essential.

\section{Limitation}
\label{sec:limitation}

Our work has two main limitations. First, we observe a limitation in temporal reasoning. Performance on Temporal Verification declines after XVR training, and models show minimal improvements on tasks involving dynamic camera movements. XVR emphasizes geometric consistency across static multi-view configurations, which reduces sensitivity to temporal dynamics. This trade-off improves structural stability across views at the cost of temporal flexibility. Future work could extend cross-view relation reasoning to explicitly incorporate temporal relationships, enabling models to understand both static spatial configurations and dynamic camera movements.

Second, our VLA transfer evaluation is conducted only in a simulation environment. While simulation provides controlled conditions for systematic analysis, it cannot fully capture the complexities of physical execution.
Extending XVR-trained models to real robot platforms would offer a more comprehensive assessment of how cross-view relation reasoning transfers to real-world manipulation, and we view this as an important direction for future work.

{
    \small
    \section*{Acknowledgments} 
    This work was supported by Institute for Information \& communications Technology Planning \& Evaluation(IITP) grant funded by the Korea government(MSIT) (RS-2019-II190075, Artificial Intelligence Graduate School Program(KAIST)); and by the Institute of Information \& Communications Technology Planning \& Evaluation(IITP) grant funded by the Korea government(MSIT) (RS-2025-02304967, AI Star Fellowship(KAIST)). This research was also conducted as part of the Sovereign AI Foundation Model Project(Data Track), organized by the Ministry of Science and ICT(MSIT) and supported by the National Information Society Agency(NIA), S.Korea. (Grant No. 2026-AIData-WII01).
}

{
    \small
    \bibliographystyle{ieeenat_fullname}
    \bibliography{main}
}

\onecolumn
{
    \centering
    \Large
    \textbf{Learning Multi-View Spatial Reasoning from Cross-View Relations}\\
    \vspace{0.5em}Supplementary Material \\
    \vspace{1.0em}
}
\setcounter{page}{1}

\renewcommand{\thesection}{\arabic{section}}
\setcounter{section}{6}

\section{Outline}

This supplementary material provides additional technical details, experimental analysis, and formal definitions omitted from the main text due to space constraints. The appendices are organized as follows:

\begin{itemize}
    \item Appendix~\ref{app:unified_formulation}: Formal mathematical definitions of the three task categories and the question-answer generation framework.
    \item Appendix~\ref{app:generation_pipeline}: Detailed task generation pipeline covering geometry-based and metadata-based generation methods.
    \item Appendix~\ref{app:temporal_filtering}: Filtering methodology for Temporal Verification.
    \item Appendix~\ref{app:dataset_statistics}: Dataset statistics including source data composition and XVR distribution analysis.
    \item Appendix~\ref{app:XVR_eval_details}: Additional analysis of XVR-Eval benchmark.
    \item Appendix~\ref{app:ood_analysis}: Task-by-task performance analysis on external benchmarks (MindCube-Tiny and RoboSpatial-Home).
    \item Appendix~\ref{app:additional_experiments}: Additional experimental results on model and data scaling.
    \item Appendix~\ref{app:td}: Training hyperparameters and procedures for VLM fine-tuning and VLA policy training.
    \item Appendix~\ref{app:qualitative}: Qualitative examples of XVR tasks.
\end{itemize}

\section{Formal Task Definitions}
\label{app:unified_formulation}

We provide formal mathematical definitions for XVR's three task categories. While Section~\ref{sec:dataset} introduces these categories conceptually, this appendix formalizes the reasoning objectives and generation framework underlying each task. Table~\ref{tab:reasoning-summary} shows how the general formulation instantiates into eight specific tasks.

\subsection{Category}
\label{app:category_definitions}

We formalize cross-view relations as a general relation between multi-view observations. 
Given a set of images $\mathcal{I} = \{I_i\}$ captured from different viewpoints,
each reasoning objective is represented by a relation function:
\begin{equation}
g_\text{task}(\mathcal{I}_r, \mathcal{I}_j, e_j)
=
\begin{cases}
c(\cdot) & \text{for Correspondence}\\[2pt]
v(\cdot) & \text{for Verification}\\[2pt]
l(\cdot) & \text{for Localization}.
\end{cases}
\end{equation}
Here, $e_j$ denotes an optional element associated with view $I_j$ 
(e.g., a point or direction), and the functions 
$c(\cdot)$, $v(\cdot)$, and $l(\cdot)$ capture complementary notions 
of geometric relationships across views.

We define three representative constraints corresponding to the three task categories:

\paragraph{(1) Correspondence:}
\begin{equation}
e^\star = \arg\max_{e \in E_j} c(\mathcal{I}_r, I_j, e)
\end{equation}
This objective finds the element $e^\star$ (e.g., a point or direction) in $I_j$ that best corresponds to the same physical entity observed in $\mathcal{I}_r$. 

\paragraph{(2) Verification:}
\begin{equation}
j^\star = \arg\min_{j} v(\mathcal{I}; \Theta)
\end{equation}
where $\Theta$ denotes the evaluation criterion (e.g., spatial consistency or temporal alignment). It identifies the view that shows the lowest consistency with the cross-view relations established by $\mathcal{I}$.

\paragraph{(3) Localization:}
\begin{equation}
j^\star = \arg\max_{j} l(\mathcal{I}; \phi)
\end{equation}
where $\phi$ represents a spatial or semantic condition (e.g., ``left of the reference camera'' or a textual description).
This constraint selects the view $I_{j^\star}$ that best satisfies the given condition.

\subsection{Question-Answer Generation}
\label{app:qa_generation}

The QA generation pipeline instantiates these formal definitions into concrete question-answer pairs from multi-view data.
Formally, each generation instance is defined as:
\begin{equation}
\label{eq:qa_framework}
\mathcal{G}:\;(\mathcal{I},\, \mathcal{P},\, \mathcal{X},\, \mathcal{T},\, \mathcal{M})
\!\rightarrow\! (\mathcal{Q},\, \mathcal{A})
\end{equation}
where 
$\mathcal{P}=\{(R_i, t_i)\}$ represents camera parameters (intrinsics/extrinsics), 
$\mathcal{X}$ denotes geometric structure such as 3D point clouds,
$\mathcal{T}=\{t_i\}$ provides temporal indices for synchronization across frames,
and $\mathcal{M}$ contains auxiliary metadata (e.g., robot poses or camera identifiers). 

Each QA pair $(\mathcal{Q}, \mathcal{A})$ is derived from a rule-based template that maps these multimodal inputs 
into reasoning objectives across the three categories: correspondence, verification, and localization.
Specifically, $\mathcal{Q}$ specifies a relational query following the prototypes in Table~\ref{tab:reasoning-summary},
and $\mathcal{A}$ provides the correct answer identifying either a view or an element within a view.


\begin{table*}[t]
\centering
\scriptsize
\caption{
Taxonomy of cross-view reasoning tasks, summarized with their formal definitions and representative QA prototypes.
}
\label{tab:qa-task}
\setlength{\tabcolsep}{4pt}
\renewcommand{\arraystretch}{2.0}
\begin{tabular}{
  @{}>{\raggedright\arraybackslash}p{3.8cm}
  >{\raggedright\arraybackslash}p{4.3cm}
  >{\raggedright\arraybackslash}p{7.0cm}@{} 
}
\toprule
\multicolumn{1}{c}{\textbf{Task}} & \multicolumn{1}{c}{\textbf{Formulation}} & \multicolumn{1}{c}{\textbf{Question Prototype}} \\
\midrule
\rowcolor{gray!10}\multicolumn{3}{@{}c}{\textbf{Correspondence}} \\[.25em]
Point Correspondence &
$p^\star = \arg\max_{p \in P_j} c(\mathcal{I}_r, I_j, p)$ &
\textit{Q:} Which point in $I_j$ corresponds to this point in $\mathcal{I}_r$? \\
Directional Correspondence &
$a^\star = \arg\max_{a \in A_j} c(\mathcal{I}_r, I_j, a)$&
\textit{Q:} Which arrow in $I_j$ points in the same direction as in $\mathcal{I}_r$? \\
\rowcolor{gray!10}\multicolumn{3}{@{}c}{\textbf{Verification}} \\[.25em]
Spatial Verification &
$j^\star = \arg\!\min_j v(\mathcal{I}; \mathcal{P})$ &
\textit{Q:} Which point $p$ in each image breaks spatial alignment? \\
Temporal Verification &
$j^\star = \arg\!\min_j v(\mathcal{I}; \mathcal{T})$ &
\textit{Q:} Which view was captured at a different timestamp? \\
\rowcolor{gray!10}\multicolumn{3}{@{}c}{\textbf{Localization}} \\[.25em]
Viewpoint Localization &
$j^\star = \arg\!\max_j l(\mathcal{I}; \mathcal{P})$ &
\textit{Q:} Which view corresponds to the camera at point $\mathcal{P}$? \\
Directional View Localization &
$j^\star = \arg\!\max_j l(\mathcal{I}; \text{dir})$ &
\textit{Q:} Which view lies to the right of the reference camera? \\
Cross-Scenario Localization &
$j^\star = \arg\!\max_j l(\mathcal{I}^{(1)}; \mathcal{I}^{(2)})$ &
\textit{Q:} Which camera in Scene~2 matches the viewpoint of Scene~1? \\
Language-conditioned Localization &
$j^\star = \arg\!\max_j l(\mathcal{I}; \phi_{\text{text}})$ &
\textit{Q:} Which view matches the description “wrist-mounted camera"? \\

\bottomrule
\end{tabular}
\label{tab:reasoning-summary}
\end{table*}

\section{Task Generation Pipeline}
\label{app:generation_pipeline}

\begin{figure}[t]
    \centering
    \includegraphics[width=\textwidth]{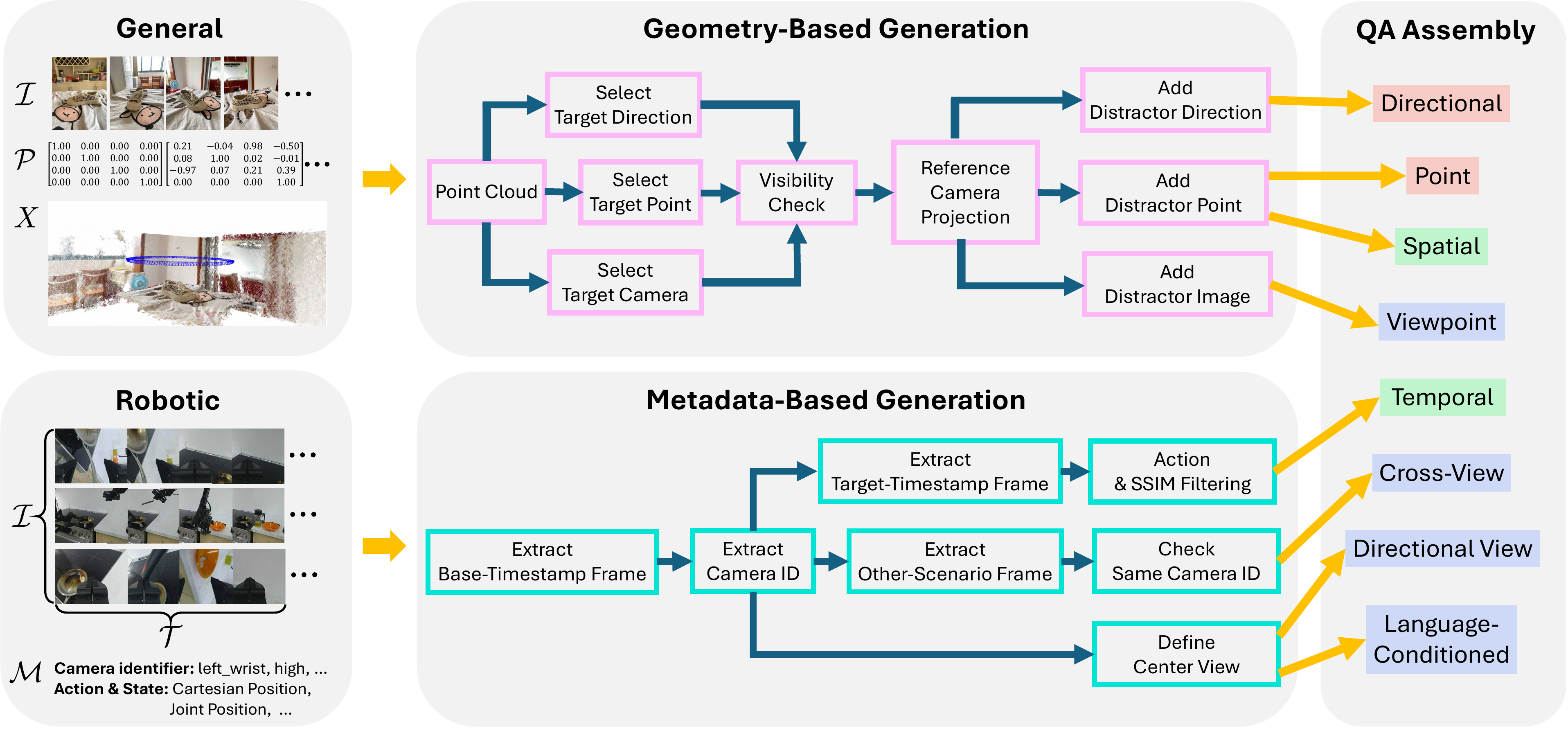}
    \caption{Task generation pipeline for XVR. The pipeline branches into geometry-based generation (top) for tasks using 3D geometric information and metadata-based generation (bottom) for tasks using trajectory annotations. Geometry-based generation processes general domain data ($\mathcal{I}, \mathcal{P}, \mathcal{X}$) through 3D-to-2D projection and visibility checking to create Point, Directional, Spatial, and Viewpoint tasks. Metadata-based generation processes robotic domain data ($\mathcal{I}, \mathcal{T}, \mathcal{M}$) through temporal and camera metadata extraction to create Temporal, Cross-View, Directional View, and Language-Conditioned tasks. Both pipelines converge at QA assembly to produce final question-answer pairs.}
    \label{fig:generation_pipeline}
\end{figure}

Figure~\ref{fig:generation_pipeline} illustrates the complete pipeline for generating XVR tasks from multi-view data. The pipeline branches based on data source characteristics: geometry-based generation for tasks leveraging explicit 3D information (general domain), and metadata-based generation for tasks utilizing trajectory annotations (robotic domain).

\subsection{Geometry-Based Generation}
\label{app:geometry_generation}

Geometry-based generation applies to tasks requiring precise 3D geometric information: Point Correspondence, Directional Correspondence, Spatial Verification, and Viewpoint Localization. The pipeline operates on general domain data with calibrated cameras and dense point clouds.

\paragraph{Input Processing.}
The pipeline receives multi-view images $\mathcal{I}$, camera parameters $\mathcal{P}$ (intrinsics and extrinsics), and 3D point clouds $\mathcal{X}$. Point clouds provide explicit geometric structure for visibility analysis and 3D-to-2D projection.

\paragraph{Target Selection.}
For correspondence tasks, the pipeline samples 3D points or directions from $\mathcal{X}$ that are visible across multiple views. For localization tasks, it samples camera positions from $\mathcal{P}$. Visibility checking ensures selected targets can be reliably projected onto reference and target views without occlusion.

\paragraph{Projection and Distractor Generation.}
Selected 3D targets are projected onto 2D image planes using camera parameters. The pipeline then generates spatially separated distractors to create challenging multiple-choice questions. For Point and Directional Correspondence, distractors are alternative points or directions within the target view. For Spatial Verification, one view receives an intentionally inconsistent point. For Viewpoint Localization, distractor images come from other camera positions.

\subsection{Metadata-Based Generation}
\label{app:metadata_generation}

Metadata-based generation applies to tasks utilizing trajectory information: Temporal Verification, Directional View Localization, Cross-Scenario Localization, and Language-Conditioned Localization. The pipeline operates on robotic domain data with temporal sequences and camera metadata.

\paragraph{Base Frame and Camera Identification.}
The pipeline first extracts a base-timestamp frame and its associated camera identifier from metadata $\mathcal{M}$ (e.g., "left\_wrist", "high"). This base frame serves as the reference point for all metadata-based tasks. Camera identifiers enable matching corresponding viewpoints across different trajectories and time steps.

\paragraph{Target Frame Extraction.}
For Temporal Verification, the pipeline searches for candidate frames at different timestamps that exhibit perceptually distinguishable changes, validated through action magnitude and SSIM filtering (Appendix~\ref{app:temporal_filtering}). For Cross-Scenario Localization, it extracts frames from different trajectory scenarios and identifies those captured from the same camera identifier as the base frame.

\paragraph{Spatial and Semantic Matching.}
For Directional View Localization, the pipeline defines the base frame as the center reference and identifies which camera positions lie in specified directions (left, right, front, back) based on robot state information in $\mathcal{M}$. For Language-Conditioned Localization, it matches camera metadata against natural language descriptions to identify views satisfying textual spatial conditions (e.g., "wrist-mounted camera").

\subsection{QA Assembly}
\label{app:qa_assembly}

Both pipelines converge at the QA assembly stage, which constructs question-answer pairs following the templates in Table~\ref{tab:reasoning-summary}. Each task receives task-specific inputs from its generation pipeline and produces structured QA pairs with reference images, target options, and correct answers.

\section{Filtering for Temporal Verification}
\label{app:temporal_filtering}

Temporal Verification task requires pairs of frames with sufficient perceptual and physical differences to ensure meaningful reasoning. We employ a two-stage filtering pipeline: (1) trajectory-level pre-filtering to remove statically inactive episodes, and (2) frame-pair validation using action-based and perceptual criteria.

\subsection{Trajectory-Level Pre-filtering}

We first eliminate trajectories with minimal physical activity by computing per-episode action variance. For each trajectory $\mathcal{T} = \{a_1, a_2, \ldots, a_T\}$ where $a_t$ denotes the action at timestep $t$, we calculate:
\begin{equation}
V(\mathcal{T}) = \sum_{t=1}^{T} \lVert a_t \rVert_2
\end{equation}
Trajectories in the bottom 20\% of action variance are filtered out, as they typically represent statically held positions with insufficient motion dynamics for temporal verification task.

\subsection{Dataset-Specific Motion Statistics}
For datasets with action metadata, we compute motion statistics to establish dynamic thresholds. For each trajectory, we measure the maximum state displacement over 1-second intervals:
\begin{equation}
M_{\text{1sec}} = \max_{t=1}^{T-f} \lVert \mathbf{s}_{t+f} - \mathbf{s}_t \rVert_2
\end{equation}
where $\mathbf{s}_t$ denotes the robot state (end-effector position or joint angles) at timestep $t$, and $f$ is the control frequency corresponding to 1 second of motion. We compute percentile statistics of $M_{\text{1sec}}$ at 10\% intervals (10th, 20th, ..., 90th percentiles) across all trajectories in each dataset. The 80th percentile of this distribution, denoted as $\tau_{\text{act}}^{(d)}$ for dataset $d$, serves as the action-based threshold for frame-pair validation.

\subsection{Frame-Pair Validation}

Given a candidate frame pair $(I_r, I_t)$ at timestamps $(t_r, t_t)$, we apply two complementary filters:

\paragraph{Action Filtering.}
For datasets with action metadata, we compute the state displacement between the two frames:
\begin{equation}
\Delta s = \lVert \mathbf{s}_{t_t} - \mathbf{s}_{t_r} \rVert_2
\end{equation}
The pair is retained only if:
\begin{equation}
\Delta s > \tau_{\text{act}}^{(d)}
\end{equation}
This ensures sufficient physical motion occurred between frames.

\paragraph{SSIM Filtering.}
To prevent near-identical frames from being included, we compute the Structural Similarity Index (SSIM)~\cite{wang2004image} between grayscale versions of the two frames. Converting to grayscale emphasizes structural changes over color variations, making the filter more sensitive to meaningful geometric transformations:
\begin{equation}
\text{SSIM}(I_r, I_t) = 
\frac{(2\mu_r \mu_t + C_1)(2\sigma_{rt} + C_2)}
{(\mu_r^2 + \mu_t^2 + C_1)(\sigma_r^2 + \sigma_t^2 + C_2)}
\end{equation}
where $\mu_r$, $\mu_t$ are mean luminance values, $\sigma_r^2$, $\sigma_t^2$ are variances, $\sigma_{rt}$ is the covariance, and $C_1$, $C_2$ are stabilization constants. Frames are discarded if:
\begin{equation}
\text{SSIM}(I_r, I_t) > \tau_{\text{ssim}}
\end{equation}

\paragraph{Dataset-Specific Thresholds.}
For datasets with action metadata, we apply both filters with $\tau_{\text{ssim}} = 0.9$. Both conditions must be satisfied for a frame pair to be retained. For datasets lacking action metadata, we apply only SSIM filtering with a more stringent threshold of $\tau_{\text{ssim}} = 0.8$ to compensate for the absence of motion-based validation.

This combined filtering strategy ensures that Temporal Verification samples contain pairs of frames with both sufficient physical motion and perceptual distinctiveness, enabling models to learn meaningful temporal reasoning rather than relying on trivial visual cues.

\begin{figure}[t]
    \centering
    \includegraphics[width=\textwidth]{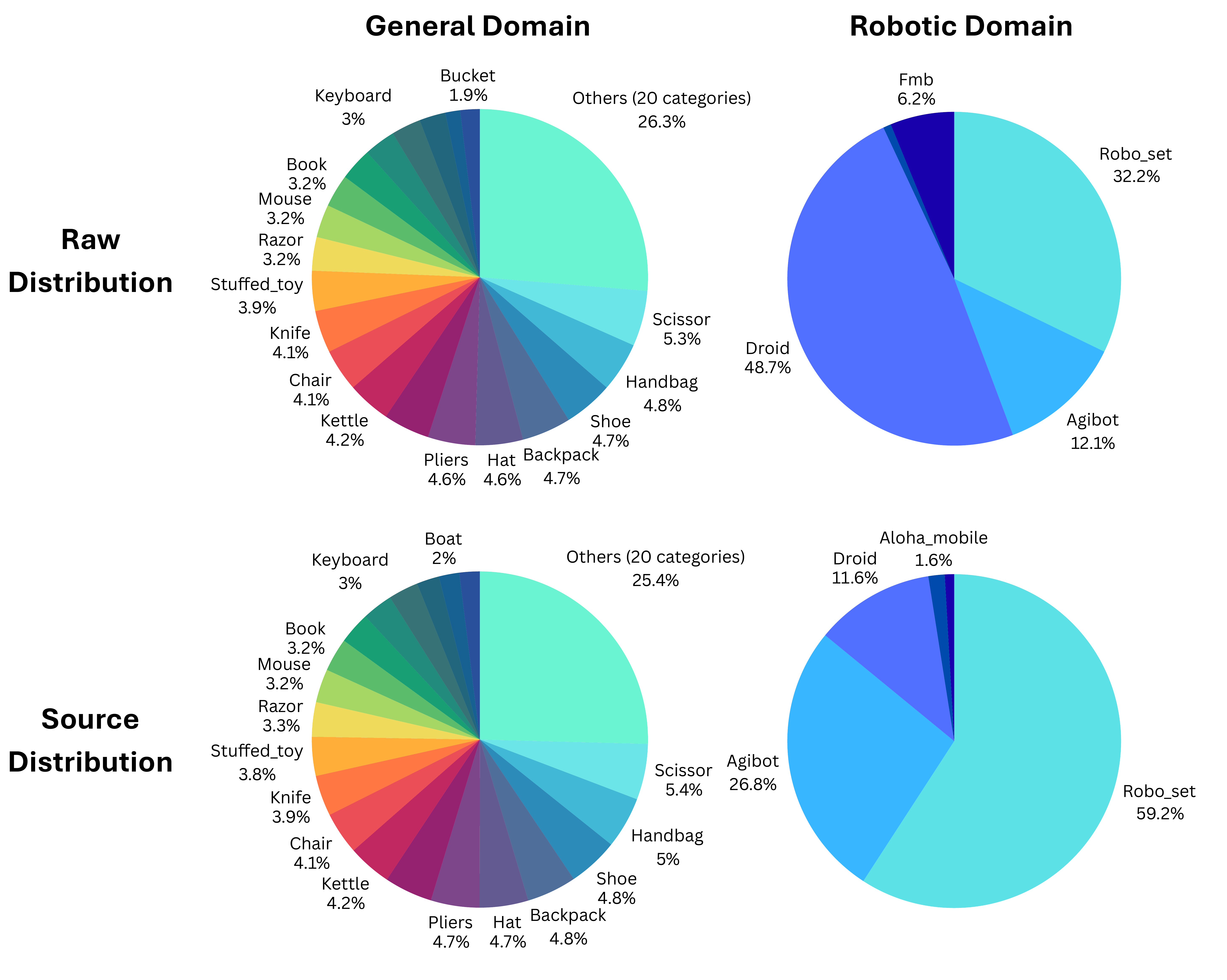}
    \caption{Source data distribution for XVR. \textbf{Top row (Raw Distribution):} Original distribution of data sources before task generation. General domain: 18,409 scenes from WildRGB-D. Robotic domain: 35,717 trajectories from DROID, RoboSet, Agibot, MobileAloha, and FMB. \textbf{Bottom row (Source Distribution):} Distribution after filtering and task generation. General domain: 51,788 samples (visibility-based filtering affects category proportions). Robotic domain: 51,788 samples (DROID and FMB limited to Temporal Verification due to inconsistent camera metadata; RoboSet and Agibot support all robotic tasks).}
    \label{fig:data_distribution}
\end{figure}

\begin{table}[t]
\centering
\caption{XVR Dataset Statistics (103,576 samples, 447,811 images)}
\label{tab:xvr_statistics}
\begin{tabular}{@{}llr@{}}
\toprule
\textbf{Category} & \textbf{Metric} & \textbf{Value} \\
\midrule
\multirow{4}{*}{Questions} 
& Avg. length (chars) & 257.7 \\
& Median length (chars) & 236.5 \\
& Min length (chars) & 25 \\
& Max length (chars) & 508 \\
\midrule
\multirow{4}{*}{Choices} 
& Avg. per question & 3.7 \\
& Median & 3.0 \\
& Min & 3 \\
& Max & 6 \\
\midrule
\multirow{4}{*}{Views per QA} 
& Avg. views per QA & 4.32 \\
& Median views & 4.0 \\
& Min views & 3 \\
& Max views & 6 \\
\midrule
\multirow{2}{*}{Image Resolution} 
& Avg. resolution & 475$\times$481 px \\
& Most common & 480$\times$640 (53.56\%) \\
\midrule
\multirow{5}{*}{Unique Resolutions}
& 480$\times$640 & 239,837 (53.56\%) \\
& 424$\times$240 & 133,501 (29.81\%) \\
& 640$\times$480 & 54,297 (12.12\%) \\
& 320$\times$180 & 18,276 (4.08\%) \\
& 256$\times$256 & 1,900 (0.42\%) \\
\midrule
\multirow{11}{*}{Answer Distribution} 
& 1 & 6,948 (6.71\%) \\
& 2 & 19,951 (19.26\%) \\
& 3 & 21,281 (20.55\%) \\
& 4 & 14,957 (14.44\%) \\
& 5 & 8,337 (8.05\%) \\
& 6 & 6,208 (5.99\%) \\
& red & 5,844 (5.64\%) \\
& blue & 5,957 (5.75\%) \\
& green & 5,756 (5.56\%) \\
& purple & 2,538 (2.45\%) \\
& yellow & 5,799 (5.60\%) \\
\bottomrule
\end{tabular}
\end{table}

\section{Dataset Statistics}
\label{app:dataset_statistics}

\subsection{Source Data Distribution}
\label{app:source_distribution}

Figure~\ref{fig:data_distribution} illustrates the distribution of source data used to construct XVR. We distinguish between raw distribution (the original data sources) and source distribution (how frequently each source contributes to XVR samples). A single raw trajectory or scene may generate multiple QA samples across different tasks, causing the source distribution to differ from the raw distribution.

\paragraph{General Domain.}
The general domain comprises diverse object categories from WildRGB-D~\cite{xia2024rgbd}. The raw distribution shows balanced coverage across object categories, with the top categories being Scissor (5.3\%), Handbag (4.8\%), Shoe (4.7\%), and Backpack (4.7\%), alongside 20 additional categories. 
The source distribution shifts due to visibility constraints. Tasks requiring 3D-to-2D projection can only be generated when geometric entities are sufficiently visible across multiple views. Scenes with limited viewpoint coverage contribute fewer samples, altering category proportions.

\paragraph{Robotic Domain.}
The robotic domain draws from DROID~\cite{khazatsky2024droid}, MobileAloha~\cite{fu2024mobile}, RoboSet~\cite{kumar2023robohive}, FMB~\cite{luo2025fmb}, and AgiBot-World~\cite{bu2025agibot}. 
The source distribution differs substantially from raw due to filtering criteria detailed in Appendix~\ref{app:temporal_filtering}. DROID and FMB contain inconsistent camera metadata, limiting their use to Temporal Verification only. RoboSet with consistent metadata supports all robotic tasks and increases proportionally in the source distribution. Additionally, trajectories satisfying multiple task requirements generate more samples, further amplifying representation differences.

\subsection{Final Dataset Statistics}
\label{app:final_stats}

Table~\ref{tab:xvr_statistics} presents comprehensive statistics for the final XVR dataset. The dataset contains 103,576 QA samples derived from 447,811 images, with an average of 4.32 views per question. Questions average 257.7 characters with 3.7 answer choices. Image resolutions vary across five distinct sizes, with 480×640 being most common (53.56\%). Answer distribution is approximately balanced across numeric choices (1-6) and color markers (red, blue, green, yellow, purple), ensuring no systematic bias toward specific answer positions.

\section{Additional XVR-Eval Analysis}
\label{app:XVR_eval_details}

\subsection{Benchmark Composition}
\label{app:xvr_eval_composition}

Table~\ref{tab:xvr_eval_distribution} presents the distribution of samples across the eight tasks in XVR-Eval. The benchmark contains 1,866 samples spanning all three task categories: Correspondence, Verification, and Localization. Random baseline accuracies vary by task structure, ranging from 20\% for tasks with average five candidate views to 50\% for binary verification tasks. The overall random baseline across all tasks is 32.64\%.

\begin{table*}[h]
\centering
\caption{XVR-Eval task distribution and random baseline accuracy.}
\label{tab:xvr_eval_distribution}
\begin{tabular}{lcc}
\toprule
Task & Samples & Random Baseline \\
\midrule
Point Correspondence & 170 (9.11\%) & 20.00\% \\
Directional Correspondence & 221 (11.84\%) & 25.00\% \\
Spatial Verification & 221 (11.84\%) & 22.22\% \\
Temporal Verification & 221 (11.84\%) & 33.33\% \\
Viewpoint Localization & 264 (14.15\%) & 33.33\% \\
Directional View Localization & 264 (14.15\%) & 50.00\% \\
Cross-Scenario Localization & 264 (14.15\%) & 33.33\% \\
Language-Conditioned Localization & 241 (12.92\%) & 50.00\% \\
\midrule
Total & 1,866 (100\%) & 32.64\% \\
\bottomrule
\end{tabular}
\end{table*}

\subsection{Out-of-Distribution Design}
\label{app:ood_design}

XVR-Eval is constructed from data sources explicitly excluded from the XVR training set to evaluate generalization capabilities. As mentioned in Section~\ref{sec:XVR_eval} of the main paper, we include two distinct held-out sources:

\paragraph{General Domain.} We use the boat category from WildRGB-D~\cite{xia2024rgbd}, which was completely excluded during XVR training. As shown in Figure~\ref{fig:data_distribution}, the training set comprises 39 other object categories, while boat represents only 2\% of the general domain and were reserved for evaluation. This tests whether models can transfer cross-view spatial reasoning learned from objects with structured shapes (chairs, keyboards, scissors, etc.) to a distinct object category with different geometric characteristics.

\paragraph{Robotic Domain.} We include trajectories from MobileAloha~\cite{fu2024mobile}, a dataset not present in the XVR training distribution. As shown in Figure~\ref{fig:data_distribution}, the robotic training data consists of RoboSet (59.2\%), AgiBot (26.8\%), DROID (11.6\%), and FMB, while MobileAloha was reserved exclusively for evaluation. This evaluates whether cross-view relation reasoning transfers to an unseen embodiment with different hardware configurations and camera setups.

This OOD design ensures XVR-Eval measures genuine generalization rather than memorization of training distributions. The consistent improvements observed on XVR-Eval (Table~\ref{tab:sft-bench-results}) despite these distribution shifts validate that cross-view relation supervision enables models to learn transferable multi-view spatial reasoning principles.

\subsection{Qualitative Examples}
\label{app:xvr_eval_examples}

\begin{figure}[H]
    \centering
    \includegraphics[width=0.5\linewidth, height=0.64\textheight]{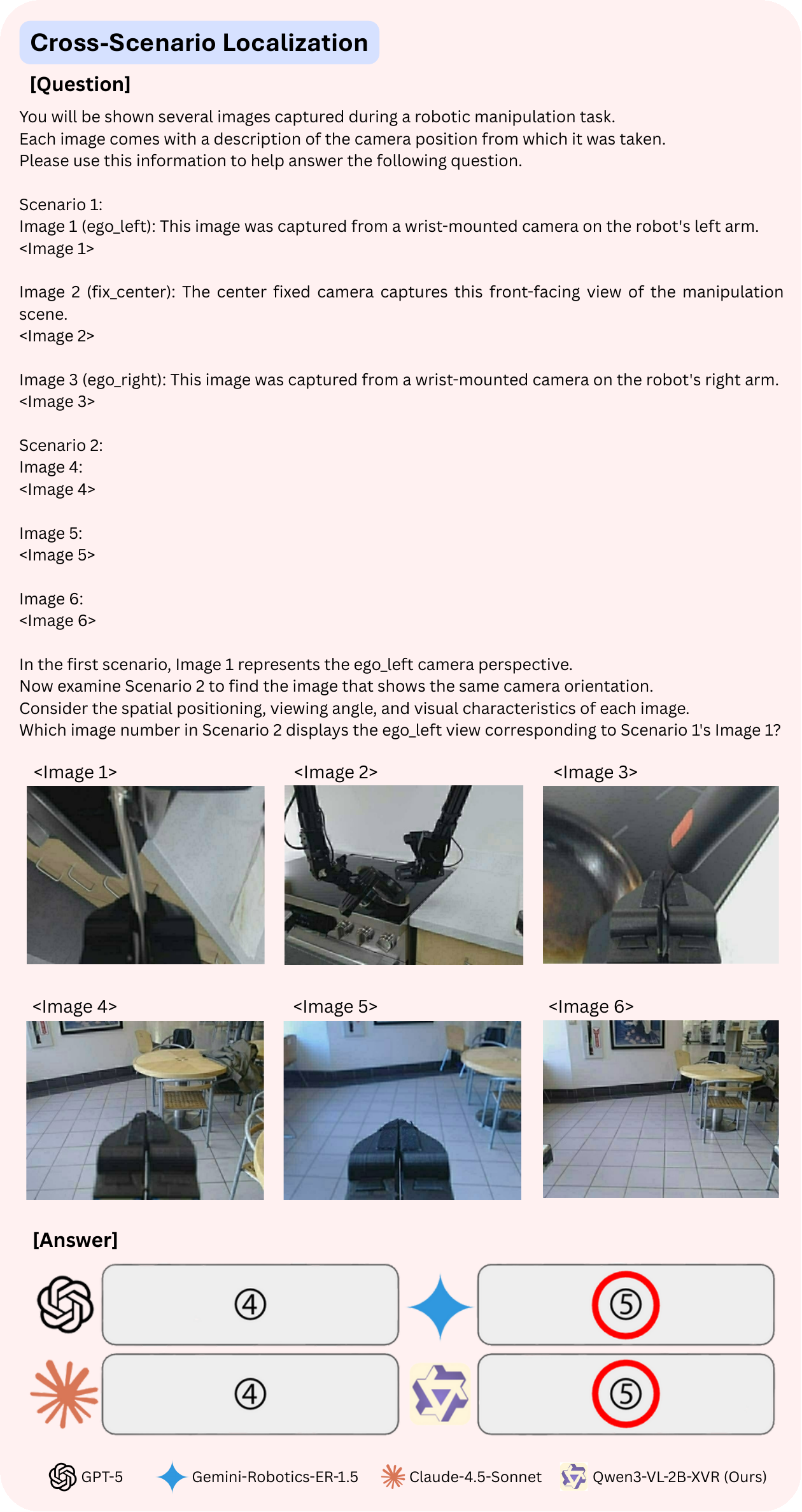}
    \caption{Cross-Scenario Localization example from XVR-Eval. The task requires identifying the corresponding viewpoint across two different scenarios. Only Qwen3-VL-2B-XVR (Ours) and Gemini-Robotics-ER-1.5 correctly predict the answer.}
    \label{fig:xvr_eval_sample1}
\end{figure}

\begin{figure}[H]
    \centering
    \includegraphics[width=0.5\linewidth, height=0.7\textheight]{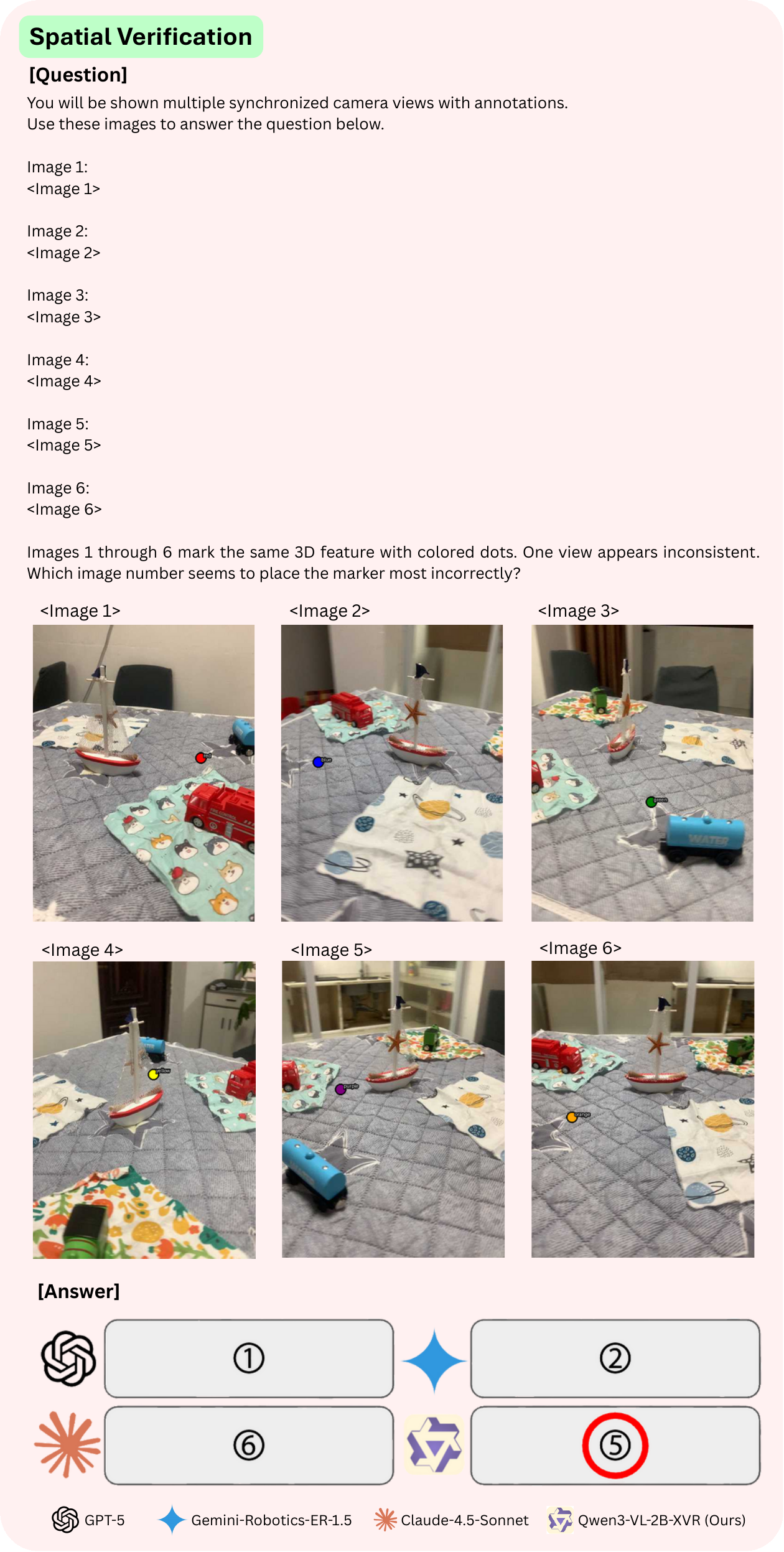}
    \caption{Spatial Verification example from XVR-Eval. The task requires identifying which view contains a marker placed at an inconsistent 3D location compared to the others. Only Qwen3-VL-2B-XVR (Ours) correctly identifies the spatially inconsistent view, demonstrating the effectiveness of explicit cross-view relation supervision.}
    \label{fig:xvr_eval_sample2}
\end{figure}

We present representative examples from XVR-Eval to illustrate the cross-view reasoning challenges and model performance patterns. Figures~\ref{fig:xvr_eval_sample1} and~\ref{fig:xvr_eval_sample2} show two samples with predictions from four models: GPT-5, Gemini-Robotics-ER-1.5, Claude-4.5-Sonnet, and our Qwen3-VL-2B-XVR.

\paragraph{Cross-Scenario Localization (Figure~\ref{fig:xvr_eval_sample1}).}
This task requires matching corresponding viewpoints across structurally similar but distinct scenes. The example demonstrates the challenge of identifying equivalent camera positions when object arrangements differ between scenarios. Among the evaluated models, only Qwen3-VL-2B-XVR and Gemini-Robotics-ER-1.5 correctly identify the matching viewpoint, demonstrating that explicit cross-view relation supervision enables robust viewpoint localization even under scene-level variations. GPT-5 and Claude-4.5-Sonnet fail on this sample, suggesting that scale alone does not guarantee consistent cross-scenario reasoning capabilities.

\paragraph{Spatial Verification (Figure~\ref{fig:xvr_eval_sample2}).}
This task requires identifying which view contains a marker at an inconsistent 3D location compared to the others. The example presents multiple views where one view shows a marker placed at a spatially inconsistent position that violates cross-view geometric constraints. Only Qwen3-VL-2B-XVR successfully identifies the inconsistent view, while all other models fail. This demonstrates that XVR training enables precise geometric coherence checking across views—a capability that closed-source models struggle with despite their larger scale.

These examples validate the effectiveness of explicit cross-view relation supervision. While closed-source models show inconsistent performance, XVR-trained models demonstrate robust reasoning on both localization across scenarios and spatial consistency verification—the core components of cross-view relation reasoning.

\FloatBarrier

\section{External Benchmark Analysis}
\label{app:ood_analysis}

\subsection{MindCube}
\label{app:mindcube_analysis}

\paragraph{Around (+0.5pp: 34.25\% → 34.75\%).}
This task requires predicting which objects become visible after the camera rotates in a specified direction and then moves forward, combining rotation and translation transformations sequentially.
XVR does not contain examples of directional camera motion followed by translation. XVR's data consists of static multi-view observations where all cameras are fixed simultaneously. The task's assumption about sequential viewpoint changes differs fundamentally from XVR's multi-view setup, resulting in minimal transfer.

\paragraph{Rotation (+0.5pp: 27.0\% → 27.5\%).}
This task provides images captured as a camera rotates 360° around a central point and requires reasoning about the complete circular spatial layout.
While this task involves multi-view reasoning, the camera configuration is outside-looking-inward, which is completely absent from XVR's training data. XVR focuses on inside-looking-outward setups (objects in front of cameras) from both general domain scenes and robotic manipulation. This distribution mismatch limits transfer despite both tasks requiring discrete multi-view understanding.

\paragraph{Among (+7.0pp: 32.5\% → 39.5\%).}
This task provides disjoint camera views where objects may be occluded or partially visible, and requires localizing a target object's position relative to other objects by integrating information across views.
The substantial improvement demonstrates that XVR successfully teaches models to understand relationships between spatial information and camera viewpoints. XVR's three task categories collectively train models to establish geometric relationships between views, verify spatial consistency, and reason about camera-relative positions. Although XVR uses geometric points rather than objects as training targets, this learned capability to reason about how spatial information appears across different viewpoints generalizes to object-level localization. 

\subsection{RoboSpatial-Home}
\label{app:robospatial_analysis}

\paragraph{Compatibility (+7.62pp: 49.52\% → 57.14\%).}
This task provides a single-view image and requires determining whether a target object can physically fit into specified empty spaces, evaluating 3D size, shape, and spatial constraints from 2D observations.
This represents the largest improvement across all external benchmarks. We attribute this to XVR creating 3D-aware representations through multi-view training. To establish correspondence across multiple 2D projections, models must learn to reason about underlying 3D structure, including spatial extent, depth relationships, and geometric constraints. This 3D reasoning capability, learned from multi-view supervision, transfers to single-view 3D tasks. The model can evaluate spatial fit and identify empty spaces from single views by applying the same geometric reasoning developed for multi-view consistency.

\paragraph{Configuration (+1.80pp: 73.17\% → 74.8\%).}
This task evaluates understanding of object-object spatial relations through classifying relations (on, in, next to), reasoning about composed relations, and localizing objects from spatial descriptions.
XVR does not contain supervision on object-object spatial relationships. The training focuses on view-view correspondence and geometric consistency rather than semantic relations between objects within scenes. The modest improvement likely reflects general spatial reasoning transfer, though the high baseline limits potential gains. The lack of explicit object-relation supervision in XVR explains why this task shows smaller improvements compared to Compatibility.

\FloatBarrier

\section{Additional Experimental Results}
\label{app:additional_experiments}

\begin{figure*}[t]
    \centering
    \includegraphics[width=1.0\linewidth]{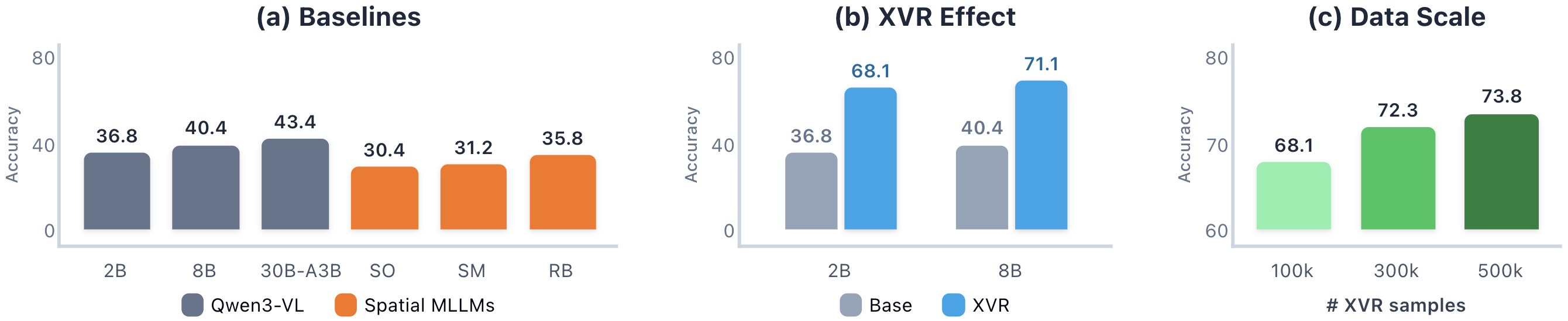}
    \caption{\textbf{Additional experiments on XVR-Eval.} (a) Baseline evaluation across Qwen3-VL model scales (2B, 8B, 30B-A3B) and spatial reasoning models: SpatialOM-3B (SO), SpatialMLLM-3B (SM), and RoboBrain2.0-32B (RB). (b) Impact of XVR training on Qwen3-VL-2B and 8B. (c) Scaling behavior of XVR training data (100k, 300k, 500k samples) on Qwen3-VL-2B.}
    \label{fig:additional_eval}
\end{figure*}

Figure~\ref{fig:additional_eval} presents additional experiments on XVR-Eval, including baseline comparisons across model scales and the impact of XVR training on model and data scaling.

\paragraph{Spatial Reasoning Baselines (a).}
We compare Qwen3-VL across model scales (2B, 8B, 30B-A3B) against models specifically designed for spatial reasoning: SpatialOM-3B (SO, 30.4\%), SpatialMLLM-3B (SM, 31.2\%), and RoboBrain2.0-32B (RB, 35.8\%). Despite being purpose-built for spatial tasks, these models underperform even the smallest general-purpose Qwen3-VL-2B (36.8\%), suggesting that cross-view spatial reasoning is not adequately covered by existing spatial reasoning datasets. Notably, RoboBrain2.0-32B, despite its 32B parameters, scores only 35.8\% overall—below Qwen3-VL-2B on most tasks including Point Correspondence (31.4\% vs. 46.6\%) and Cross-Scenario Localization (41.2\% vs. 41.6\%).

\paragraph{Model Scaling (b).}
XVR training benefits extend beyond the 2B scale. Qwen3-VL-2B improves from 36.8\% to 68.1\% (+31.3pp) after XVR fine-tuning, and Qwen3-VL-8B similarly improves from 40.4\% to 71.1\% (+30.7pp), demonstrating consistent gains across model scales. Task-level improvements are also consistent: Point Correspondence improves from 65.5\% to 98.1\% (+32.6pp) on 8B, and Spatial Verification improves from 23.9\% to 91.3\% (+67.4pp), confirming that explicit cross-view supervision is broadly effective regardless of model capacity.

\paragraph{Data Scaling (c).}
We investigate scaling beyond 100K samples by generating additional data from the same source distribution. Overall performance improves consistently from 68.1\% (100K) to 72.3\% (300K) to 73.8\% (500K). Task-level gains are particularly notable on geometry-intensive tasks: Point Correspondence reaches 98.9\% at 500K, and Viewpoint Localization improves from 57.7\% to 88.0\%, confirming that cross-view reasoning benefits from larger training sets without saturating at 100K.

\FloatBarrier

\section{Training Details}
\label{app:td}
This section provides complete implementation details for reproducing our experimental results. We describe two training stages: (1) fine-tuning the vision-language model on XVR to acquire cross-view spatial reasoning capabilities (\S\ref{app:vlm-finetuning}), and (2) extending the XVR-trained VLM into a Vision-Language-Action model for embodied manipulation tasks (\S\ref{app:vla-extension}). 

The first stage trains Qwen3-VL-2B-Instruct on the 100K XVR dataset through full-parameter supervised fine-tuning, producing the Qwen3-VL-2B-XVR backbone used in all main paper experiments. The second stage integrates this XVR-trained backbone into a VLA architecture by adding a diffusion-based action head following the GR00T-N1.5 design. We evaluate the resulting policies on RoboCasa simulation tasks to assess the transfer of learned spatial representations to robotic control.
All experiments are conducted on NVIDIA H100 GPUs using distributed training frameworks. We provide full hyperparameter specifications below to ensure reproducibility.

\subsection{Fine-Tuning VLM on XVR}
\label{app:vlm-finetuning}
To equip the base vision-language model with explicit cross-view spatial reasoning capabilities, we fine-tuned Qwen3-VL-2B-Instruct on the full XVR dataset. All experiments were conducted using \texttt{HuggingFace Accelerate}, which internally adopts Distributed Data Parallel (DDP) across a single node with 8$\times$~NVIDIA~H100 GPUs. We perform full-parameter supervised fine-tuning and enabled both gradient checkpointing and gradient accumulation to efficiently process variable-resolution multi-view inputs.

The XVR dataset contains 103{,}576 multi-view QA samples. While the image resolutions vary, the average resolution is 475$\times$481, and the most common resolution is 480$\times$640 (53.6\% of all samples). A unified training script was used across all runs, with only the dataset configuration differing between experiments to ensure comparability.
The full set of optimization hyperparameters is summarized in Table~\ref{tab:xvr_train_hparams}. The resulting model, denoted as \textbf{Qwen3-VL-2B-XVR}, serves as the spatial reasoning backbone for all VLM and VLA experiments presented in the main paper.

\begin{table}[h!]
\centering
\caption{Training hyperparameters for XVR fine-tuning of Qwen3-VL-2B-Instruct.}
\label{tab:xvr_train_hparams}
\vspace{0.2cm}
\begin{tabular}{l c}
\toprule
\textbf{Parameter} & \textbf{Value} \\
\midrule
Model & Qwen3-VL-2B-Instruct \\
Dataset size & 103{,}576 QA pairs \\
Epochs & 3 \\
Learning rate & 5e-5 \\
Scheduler & Cosine \\
Fine-tuning type & Full-parameter \\
Global batch size & 256 \\
Per-device batch size & 4 \\
Gradient accumulation steps & 8 \\
GPUs used & 8$\times$~NVIDIA H100 (1 node) \\
Distributed strategy & Accelerate (DDP) \\
Gradient checkpointing & True \\
Max grad norm & 1 \\
Precision & BF16 \\
Optimizer & AdamW \\
Warmup ratio & 0.05 \\
Average image resolution & 475$\times$481 \\
Most common resolution & 480$\times$640 (53.6\%) \\
\bottomrule
\end{tabular}
\end{table}
\label{app:training_details}

\subsection{VLA Policy Training with XVR-Trained VLM}
\label{app:vla-extension}
To evaluate whether XVR supervision can improve downstream robotic control, we train a Vision-Language-Action (VLA) policy using the Qwen3-VL-2B-XVR as the perceptual backbone. Our architecture follows the design of Isaac GR00T N1.5: visual observations and language instructions are processed through the VLM, while robot proprioceptive states and noised actions are fed as inputs to a DiT-based action head. The VLM output embedding (taken from the 12th transformer layer) is used as a conditioning signal inside the DiT layers via cross-attention, allowing the action transformer to attend to the XVR-encoded scene representation throughout the denoising process.

We evaluate two backbone variants: (1) replacing GR00T's original Eagle2.5 VLM with Qwen3-VL-2B-Instruct, and (2) replacing it with our \textbf{Qwen3-VL-2B-XVR}. For both variants, we freeze the language model while fine-tuning the vision encoder and training the DiT action head from scratch. All policies are trained for 60{,}000 steps on three tasks---\textit{CoffeePressButton}, \textit{TurnOffMicrowave}, and \textit{PnPCabToCounter}---sourced from the \texttt{nvidia/PhysicalAI-Robotics-GR00T-X-Embodiment-Sim} dataset. All evaluations are conducted in the RoboCasa simulator, and performance is measured using success rate at the 60k-step checkpoint.

The hyperparameters used for VLA policy training are summarized in Table~\ref{tab:vla_hyperparameters}.

\begin{table}[h]
\centering
\caption{Training hyperparameters for VLA policy learning.}
\label{tab:vla_hyperparameters}
\small
\begin{tabular}{lr}
\toprule
\textbf{Hyperparameter} & \textbf{Value} \\
\midrule
Backbone VLM & Qwen3-VL-2B-Instruct / Qwen3-VL-2B-XVR \\
Frozen components & Language model (LLM) \\
Fine-tuned components & Vision encoder, Projector, DiT action head \\
Action head & DiT (trained from scratch) \\
VLM feature layer used & Transformer layer 12 \\
Tasks & CoffeePressButton, TurnOffMicrowave, PnPCabToCounter \\
Dataset & \texttt{nvidia/PhysicalAI-Robotics-GR00T-X-Embodiment-Sim} \\
Training steps & 60,000 \\
Batch size (per GPU) & 16 \\
Total batch size & 128 (8 GPUs) \\
Learning rate & $1 \times 10^{-4}$ \\
Weight decay & $1 \times 10^{-5}$ \\
Optimizer & AdamW ($\beta_1{=}0.95$, $\beta_2{=}0.999$) \\
Scheduler & Cosine \\
Warmup ratio & 0.05 \\
Mixed precision & BF16 \\
Hardware & 8$\times$ NVIDIA H100 \\
Evaluation metric & Success rate at 60k checkpoint \\
\bottomrule
\end{tabular}
\end{table}

\section{XVR Examples}
\label{app:qualitative}

\begin{figure}[t]
    \centering
    \includegraphics[width=\linewidth, height=0.95\textheight, keepaspectratio]{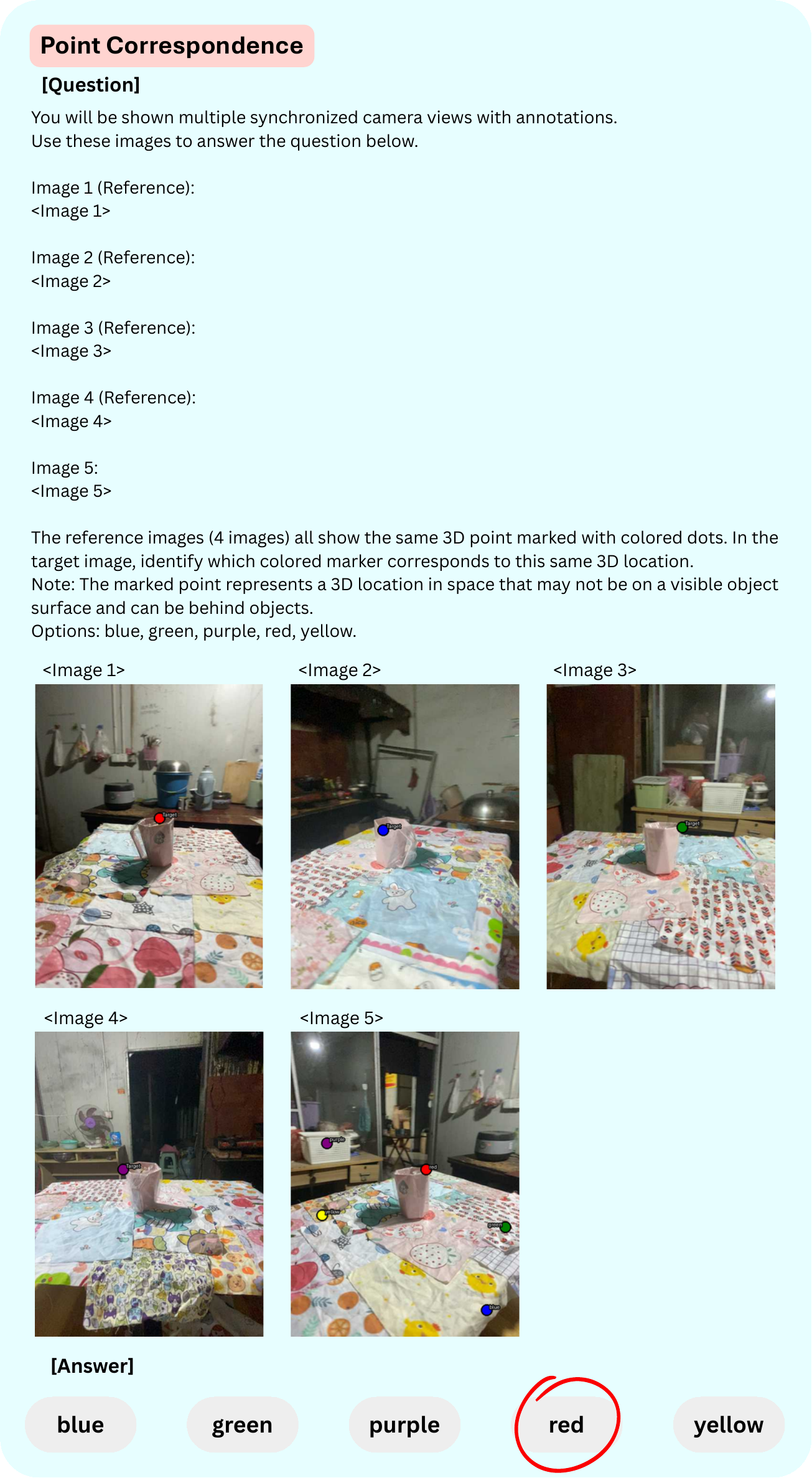}
    \caption{Point Correspondence task example. The question asks which point in the target image corresponds to a marked point in the reference images.}
    \label{fig:example_pc}
\end{figure}

\begin{figure}[t]
    \centering
    \includegraphics[width=\linewidth, keepaspectratio]{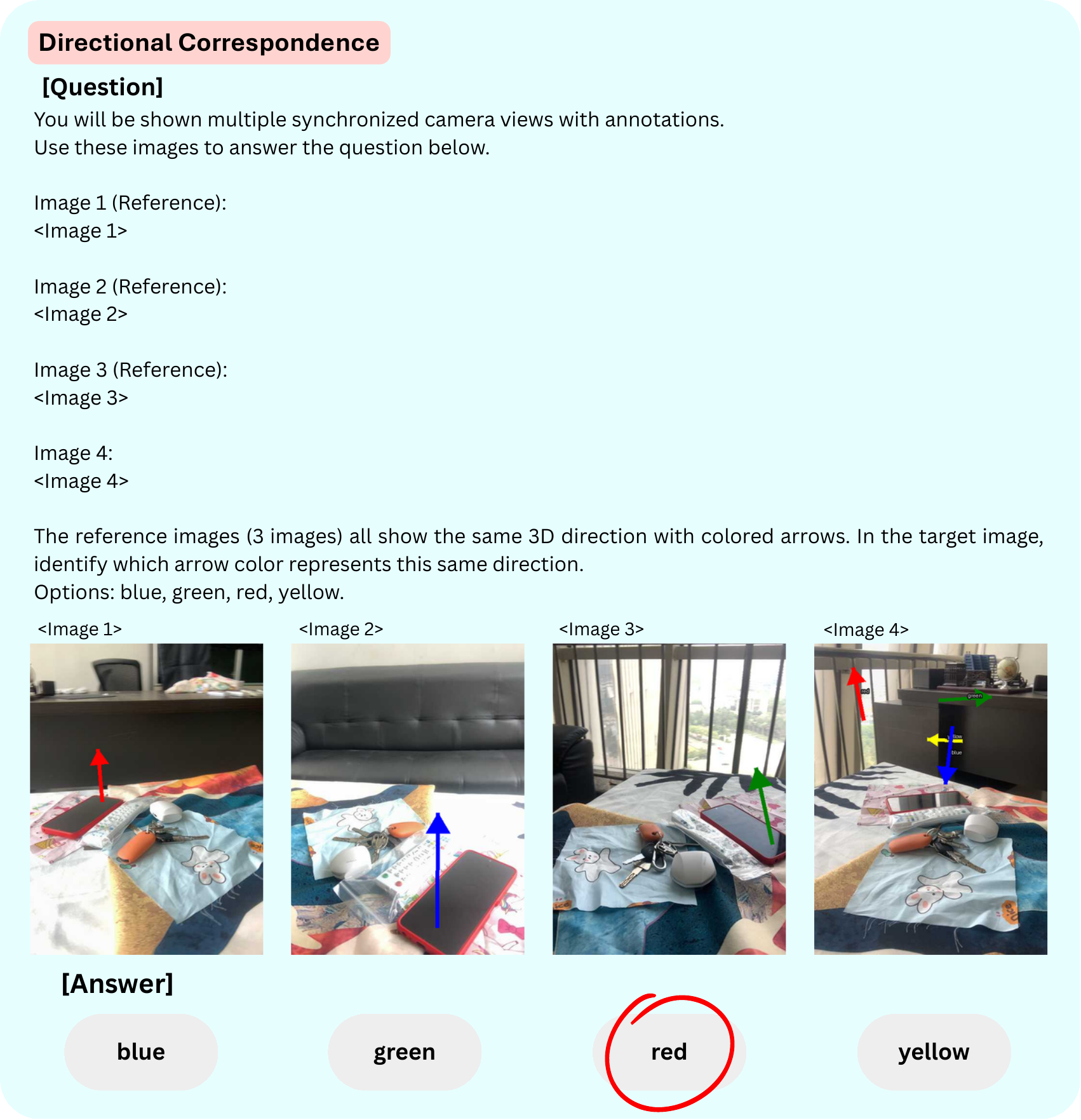}
    \caption{Directional Correspondence task example. The question asks which arrow in the target image points in the same direction as arrows in the reference images.}
    \label{fig:example_dc}
\end{figure}

\begin{figure}[t]
    \centering
    \includegraphics[width=0.8\linewidth, keepaspectratio]{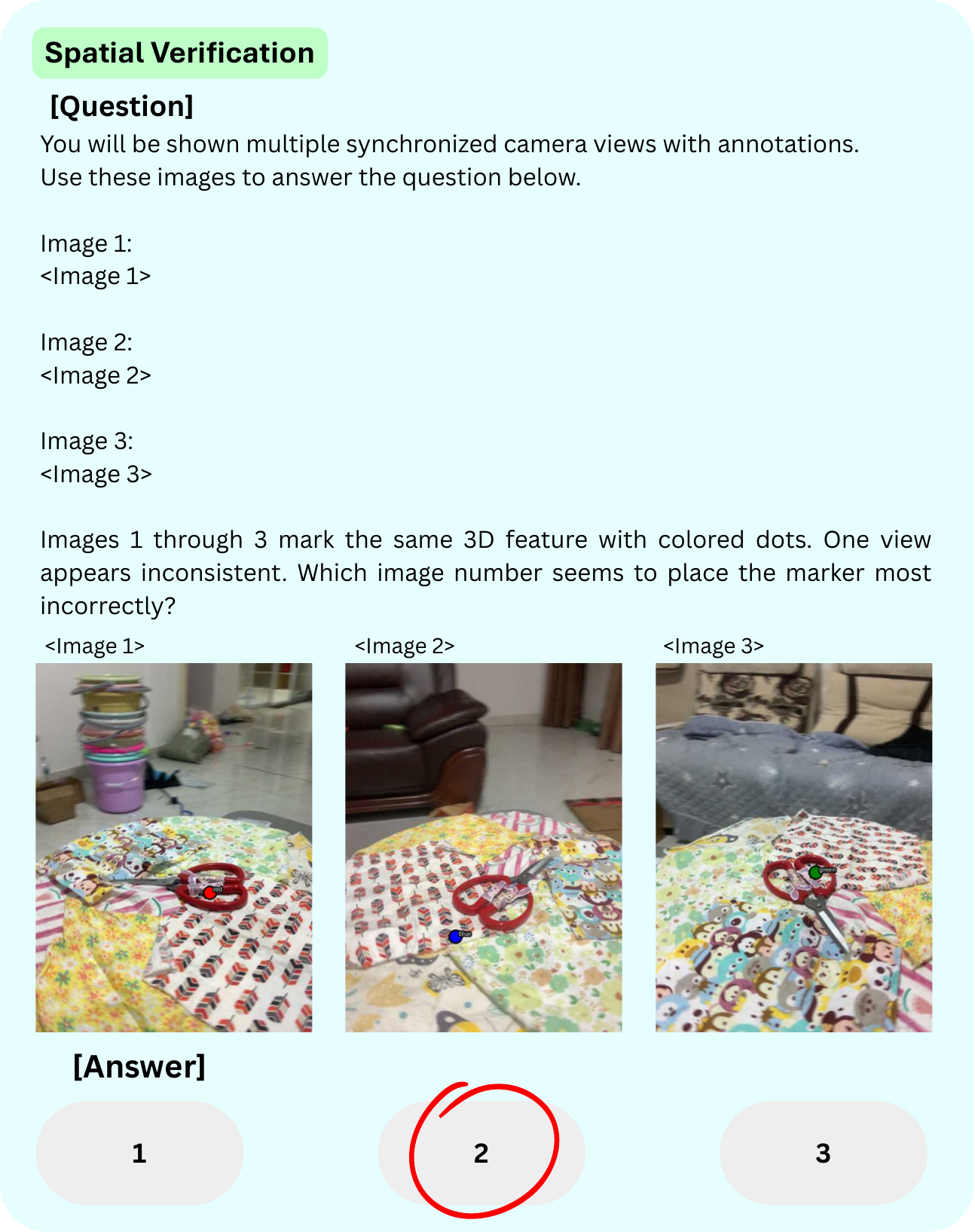}
    \caption{Spatial Verification task example. The question asks which marked point across multiple images breaks spatial alignment.}
    \label{fig:example_sv}
\end{figure}

\begin{figure}[t]
    \centering
    \includegraphics[width=\linewidth, height=0.95\textheight, keepaspectratio]{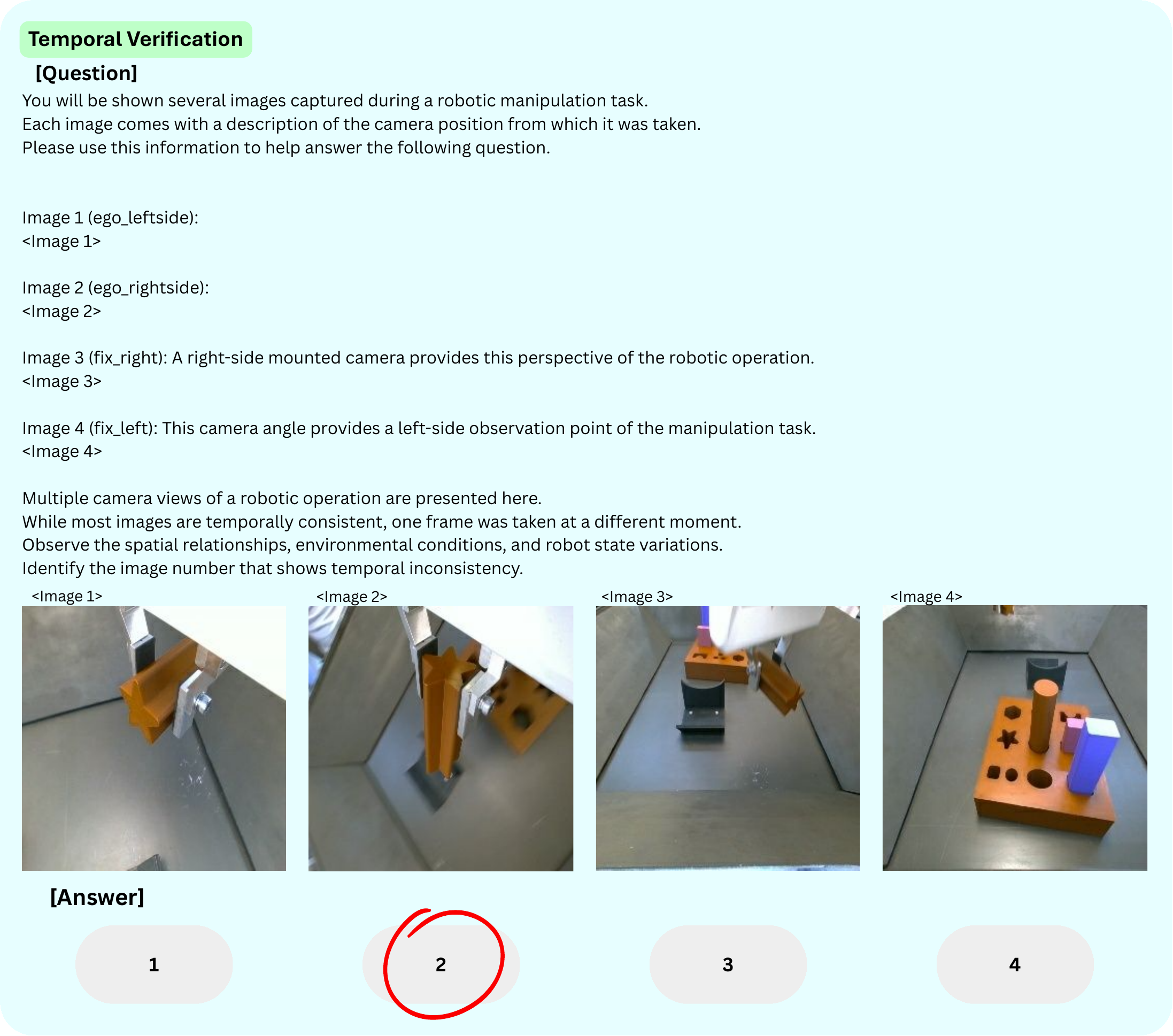}
    \caption{Temporal Verification task example. The question asks which image was captured at a different timestamp compared to others.}
    \label{fig:example_tv}
\end{figure}

\begin{figure}[t]
    \centering
    \includegraphics[width=\linewidth, height=0.95\textheight, keepaspectratio]{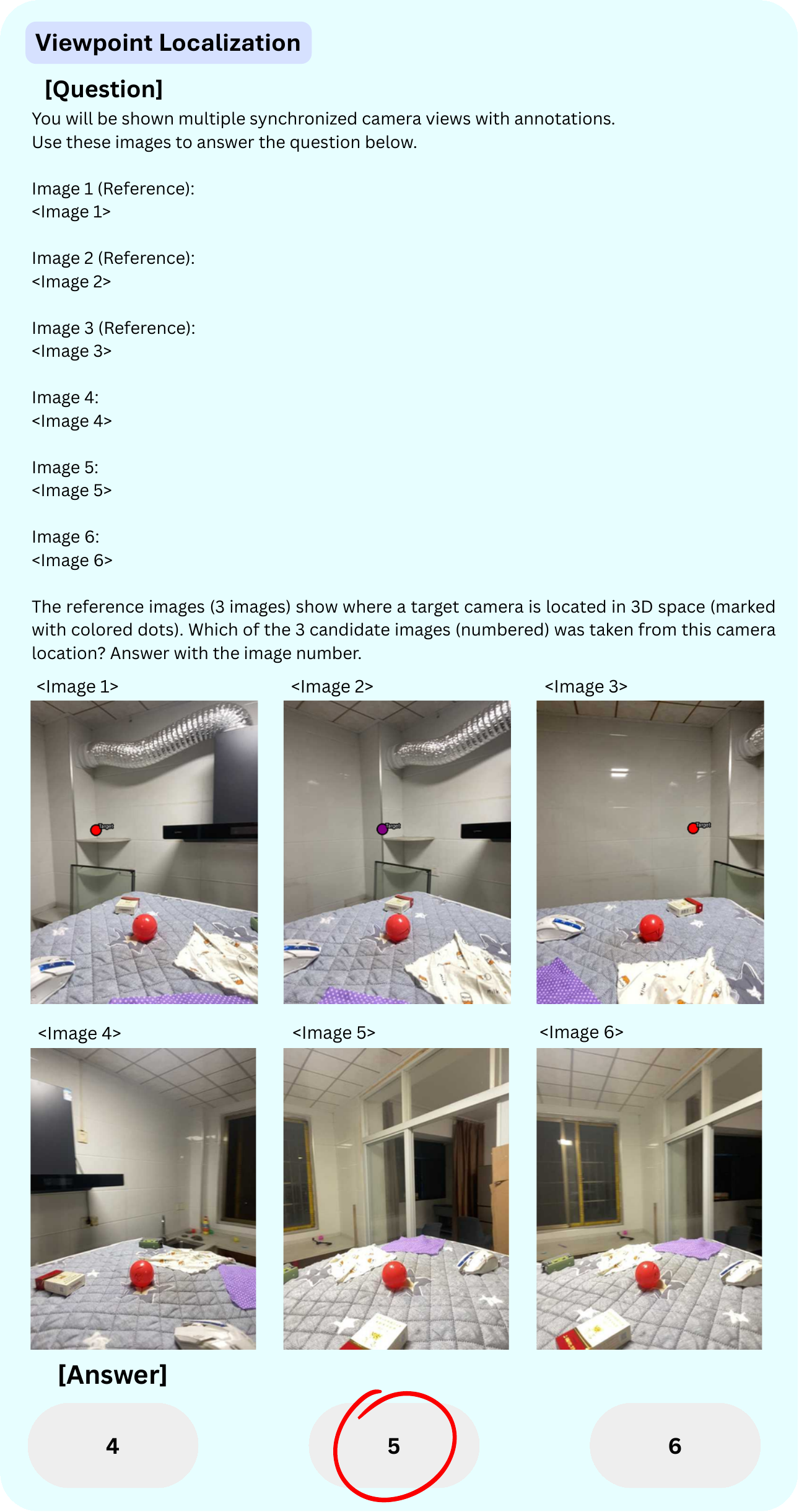}
    \caption{Viewpoint Localization task example. The question asks which camera view corresponds to a specific spatial position marked in 3D space.}
    \label{fig:example_vl}
\end{figure}

\begin{figure}[t]
    \centering
    \includegraphics[width=0.8\linewidth, keepaspectratio]{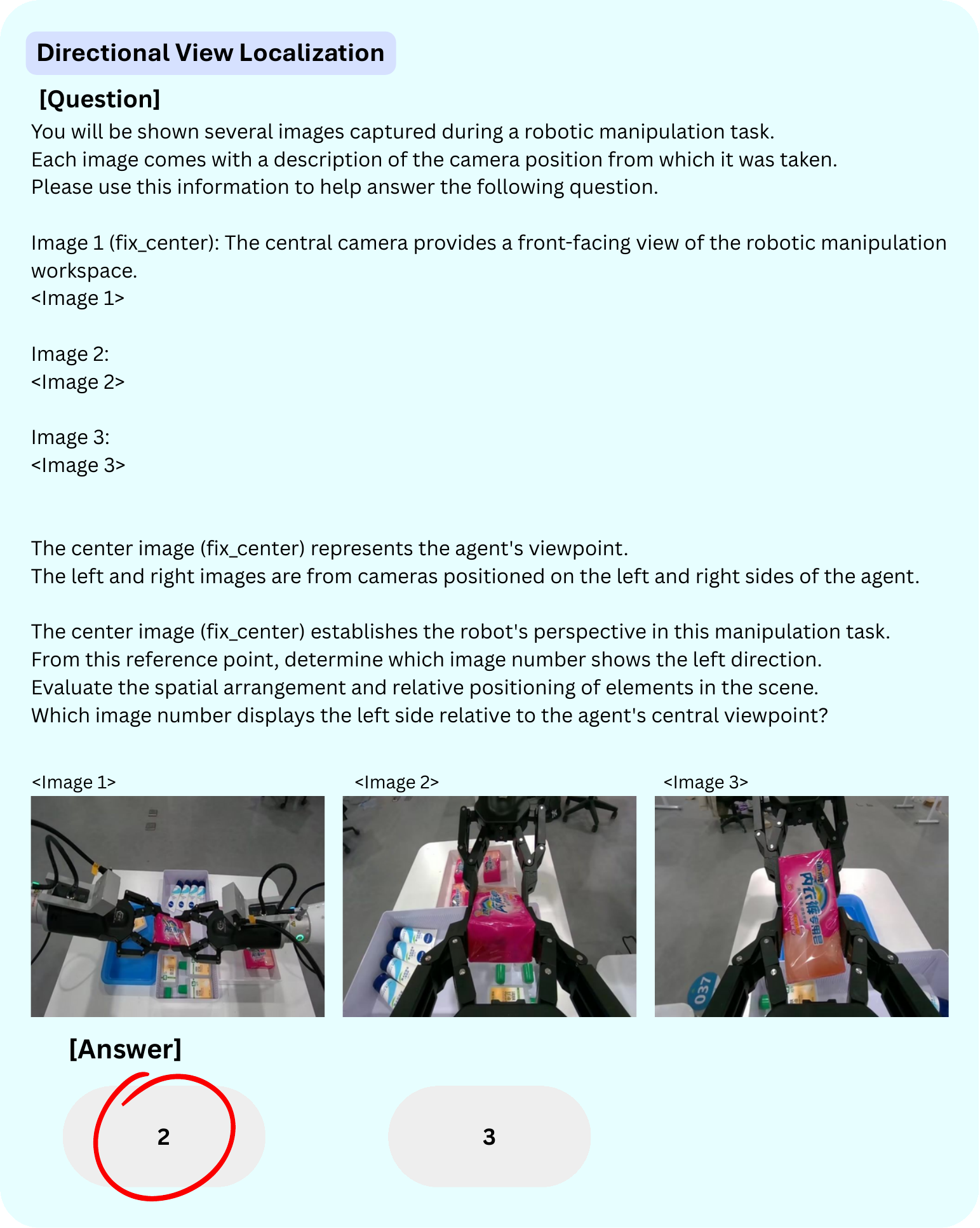}
    \caption{Directional View Localization task example. The question asks which camera view lies in a specified direction (e.g., left, right) relative to the reference camera.}
    \label{fig:example_dvl}
\end{figure}

\begin{figure}[t]
    \centering
    \includegraphics[width=\linewidth, height=0.95\textheight, keepaspectratio]{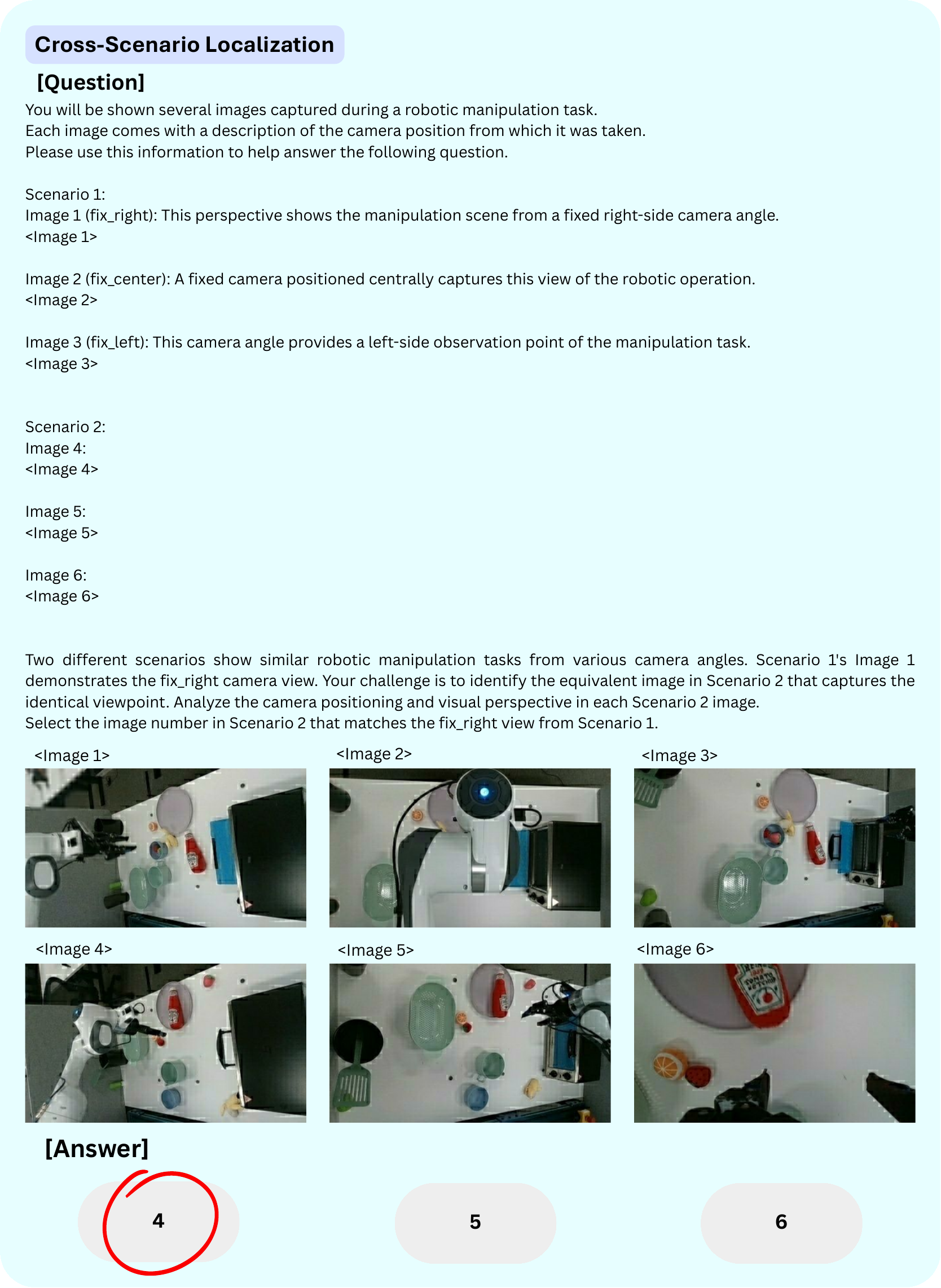}
    \caption{Cross-Scenario Localization task example. The question asks which camera view in one scenario matches the viewpoint of a reference image from another scenario.}
    \label{fig:example_csl}
\end{figure}

\begin{figure}[t]
    \centering
    \includegraphics[width=0.8\linewidth, keepaspectratio]{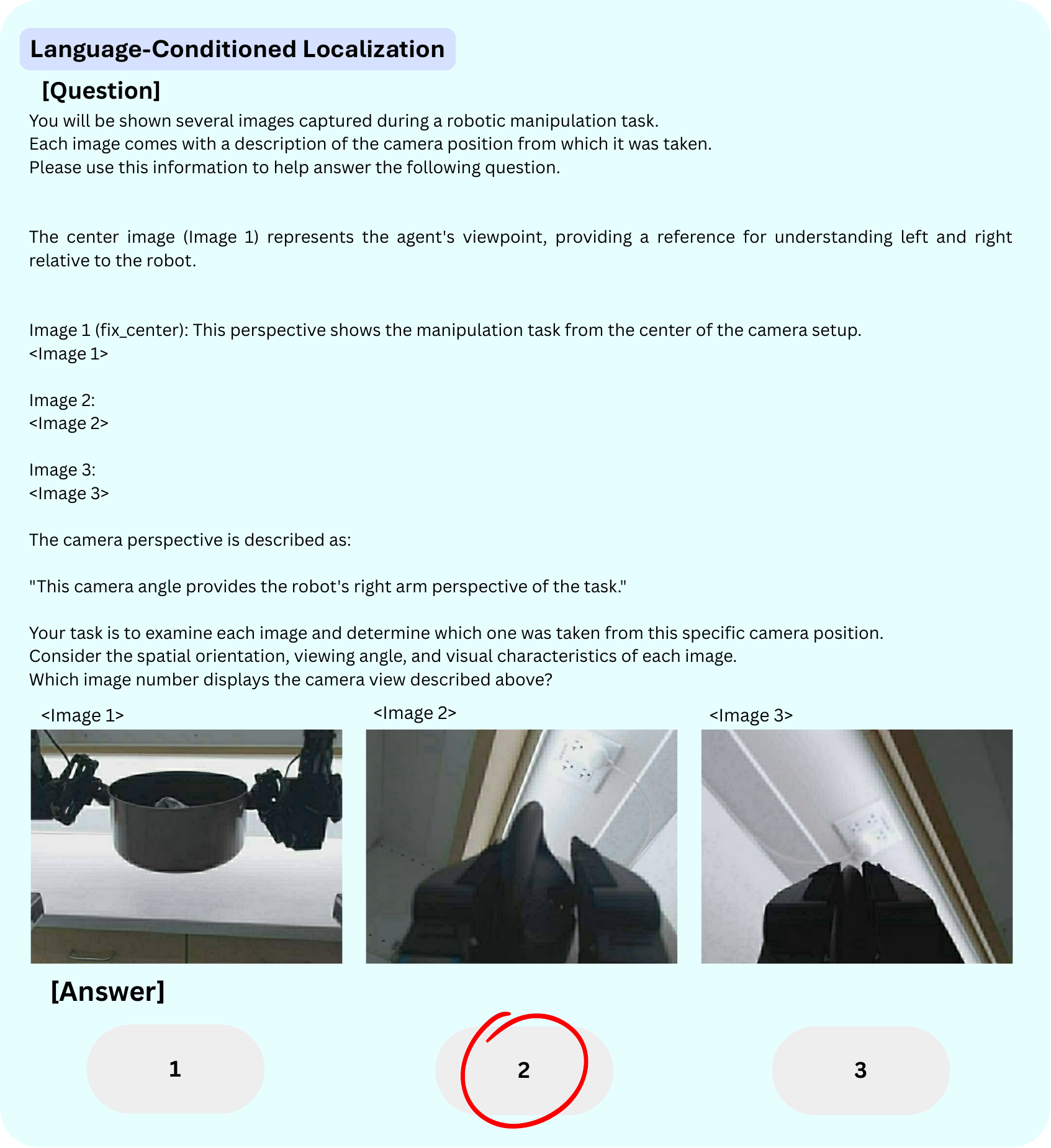}
    \caption{Language-Conditioned Localization task example. The question asks which camera view matches a natural language spatial description (e.g., "wrist-mounted camera").}
    \label{fig:example_lcl}
\end{figure}

This section provides visual examples for each of the eight tasks in XVR, illustrating the question format, reference images, and answer choices across the three task categories: Correspondence, Verification, and Localization (Figures~\ref{fig:example_pc}–\ref{fig:example_lcl}).

\end{document}